\documentclass{article}
\usepackage{microtype}
\usepackage{graphicx}
\usepackage{booktabs} % for professional tables
\usepackage{subcaption}
\usepackage{hyperref}
\usepackage{amsmath}
\usepackage{natbib}
\usepackage{cleveref}
\usepackage[accepted]{icml2020}
\icmltitlerunning{Evolutionary-Driven Reinforcement Learning}
\begin{document}
\twocolumn[
\icmltitle{Evo-RL: Evolutionary-Driven Reinforcement Learning}
\begin{icmlauthorlist}
\icmlauthor{Ahmed Hallawa}{ice,uka}
\icmlauthor{Thorsten Born}{ti}
\icmlauthor{Anke Schmeink}{ti}
\icmlauthor{Guido Dartmann}{trier}
\icmlauthor{Arne Peine}{uka}
\icmlauthor{Lukas Martin}{uka}
\icmlauthor{Giovanni Iacca}{trento}
\icmlauthor{A.E. Eiben}{amsterdam}
\icmlauthor{Gerd Ascheid}{ice}
\end{icmlauthorlist}

\icmlaffiliation{ice}{Chair for Integrated Signal Processing Systems, RWTH Aachen University}
\icmlaffiliation{uka}{Department of Intensive Care and Intermediate Care, University Hospital RWTH Aachen}
\icmlaffiliation{trier}{Research Area Distributed Systems, Trier University of Applied Sciences}
\icmlaffiliation{ti}{Research Area Information Theory and Systematic Design of Communication Systems, RWTH Aachen University}
\icmlaffiliation{amsterdam}{ Computer Science Department Vrije Universiteit Amsterdam}
\icmlaffiliation{trento}{Department of Information Engineering and Computer Science, University of Trento}

\icmlcorrespondingauthor{Ahmed Hallawa}{hallawa@ice.rwth-aachen.de}

%     ( RWTH Aachen University) <hallawa@ice.rwth-aachen.de>
%     ( RWTH Aachen University) <thorsten.born@rwth-aachen.de>
%    Anke Schmeink ( RWTH Aachen University) <Anke.Schmeink@rwth-aachen.de>
%    Guido Dartmann ( University of Applied Sciences Trier) <g.dartmann@umwelt-campus.de>
%    Arne Peine ( University Hospital RWTH Aachen) <apeine@ukaachen.de>
%    Lukas Martin ( University Hospital RWTH Aachen) <lmartin@ukaachen.de>
%    Giovanni Iacca ( University of Trento) <giovanni.iacca@unitn.it>
%    Gusz Eiben ( Vrije Universiteit Amsterdam) <a.e.eiben@vu.nl>
%    Gerd Ascheid ( RWTH Aachen University) <gerd.ascheid@ice.rwth-aachen.de>
    
\vskip 0.3in

]

\printAffiliationsAndNotice{} % otherwise use the standard text.
\begin{abstract}
In this work, we propose a novel approach for reinforcement learning driven by evolutionary computation. Our algorithm, dubbed as Evolutionary-Driven Reinforcement Learning (Evo-RL), embeds the reinforcement learning algorithm in an evolutionary cycle, where we distinctly differentiate between purely evolvable (instinctive) behaviour versus purely learnable behaviour. Furthermore, we propose that this distinction is decided by the evolutionary process, thus allowing Evo-RL to be adaptive to different environments. In addition, Evo-RL facilitates learning on environments with rewardless states, which makes it more suited for real-world problems with incomplete information. To show that Evo-RL leads to state-of-the-art performance, we present the performance of different state-of-the-art reinforcement learning algorithms when operating within Evo-RL and compare it with the case when these same algorithms are executed independently. Results show that reinforcement learning algorithms embedded within our Evo-RL approach significantly outperform the stand-alone versions of the same RL algorithms on OpenAI Gym control problems with rewardless states constrained by the same computational budget. 
\end{abstract}
\section{Introduction}

In reinforcement learning (RL), the reward function plays a pivotal role in the performance of the algorithm. However, the reward function definition is challenging especially in real world problems. This is because in many problems the designer has incomplete information regarding the problem and therefore,  defining a reward function that takes into consideration all possible states is hard. For example, in \cite{paper:AIClinician}, the authors attempted to use reinforcement learning to find the best volume therapy for patients in intensive care units. As the model targets optimizing patient mortality, the authors defined the reward based on how probabilistically, according to the patients data, a given state is close to one of two terminal states: the state of survival (positive reward) or the state of death (negative reward). Although this is a valid reward function for states close to those terminal states, it is not accurate for states that are far from both terminal states. Generally, it is hard for clinicians to quantify a reward that is valid for all possible patient states, especially if the state vector includes tens of variables. This is also valid other fields such as robotics. The question then is: Is it possible to solve a reinforcement learning problem with a reward function that is only known and valid for a few states?

Meta-heuristic algorithms such as Evolutionary Algorithms (EAs) can offer a solution to this problem. EAs is an umbrella term used to describe computer-based problem-solving systems that adopt computational models where the evolutionary processes are the main element in their design \cite{fister2015adaptation}. These algorithms are particularly good as black-box optimizers, i.e., to solve problems whose mathematical formalization is hard to produce. However, these algorithms treat the problem of finding an optimal policy of an agent interacting with a certain environment as a black-box problem, thus, not gaining any information from potential feedback signals available in the environment.

In this work, we propose a hybrid approach combining EAs and RL. Most importantly, our approach handles problems where the reward function is not available in many of the state in the state space. Of note, many previous works tried to use EA combined with RL, such as in \cite{kim2007hybrid}, \cite{hamalainen2018ppo} and \cite{koulouriotis2008reinforcement}. However, they target optimizing the RL process itself via the EA, rather than evolving an agent behaviour and embedding the learning process within it.  

Our algorithm combines evolutionary computation and RL in a single framework, where the line between which behaviour is evolvable and which one is learnable is automatically generated based on the environment properties, thus, making this approach highly adaptive. In other words, our approach does not require, at the design time, defining which part of the agent behaviour (policy) should rely on EA and which one is learned via RL, as this is also challenging in real world problems. Generally, our approach: (1) consists of an EA loop embedding RL; (2) can handle complex environments where the reward function is valid only for a limited number of states; (3) at design time, it does not require defining which part of the behaviour should be evolved and which part should be learned; (4) is agnostic with respect to the behaviour representation, and works with any reinforcement learning algorithm.

Our primary contribution is the design and implementation of an evolutionary-driven reinforcement learning algorithm. We demonstrate the algorithm on three different OpenAI gym control problems (CartPole, Acrobot and MountainCar), after transforming them into environments with rewardless states. Our evaluations show that our approach compares favorably against stand-alone RL algorithms such as Q-learning, Proximal Policy Optimization (PPO) \cite{schulman2017proximal} and Deep Q-Network \cite{mnih2015human} (DQN).

\section{Proposed Model}
We aim at handling problems with ambiguous environments where the reward function definition is hard for a wide range of states, thus, the environment presents states where the reward function is not applicable. In this section, we define the main terminology needed to explain our approach, then present the general form of our algorithm, including its design axioms.

\subsection{Terminology}
In the proposed work, we identify two types of behaviors: a purely evolved behavior, and a learnable behavior, which is driven by the agent's experience in its lifetime. The first one is dubbed as \emph{instinctive behavior}. We formally define an \textit{instinctive behaviour} as the evolved part of the agent's behaviour that is inherited from its ancestors and cannot be changed during the learning process within the lifetime of the agent. 

As for the second behavior, the learned behavior, we picture it as an extension to the evolved instinctive behavior. We define a \textit{learnable behaviour} as the behaviour learned by the agent during its lifetime, as a result of its exposure to the environment. It should be noted that the learned behaviour cannot alter the instinctive behaviour. Finally, we define the \textit{overall behaviour} as the combination of the agent's instinctive and learned behaviour, integrated together during the agent's lifetime.

Furthermore, inspired by \cite{eiben2013triangle}, we identify three states for the agent: the \emph{born} state, which means that an agent has already an instinctive behaviour, but no learned behaviour. Note that the agent in this state is not exposed to the environment yet. The second state, dubbed as \emph{mature}, means that the agent is already trained on the environment and has now an instinctive and a learned behaviour. Finally, the \emph{fertile} state, means that the agent overall behaviour is already evaluated, therefore, a score reflecting its performance relative to a pre-defined objective can be computed.

\subsection{Our Approach}
\begin{figure}[t]
        \centering
        \includegraphics[width=0.5\textwidth]{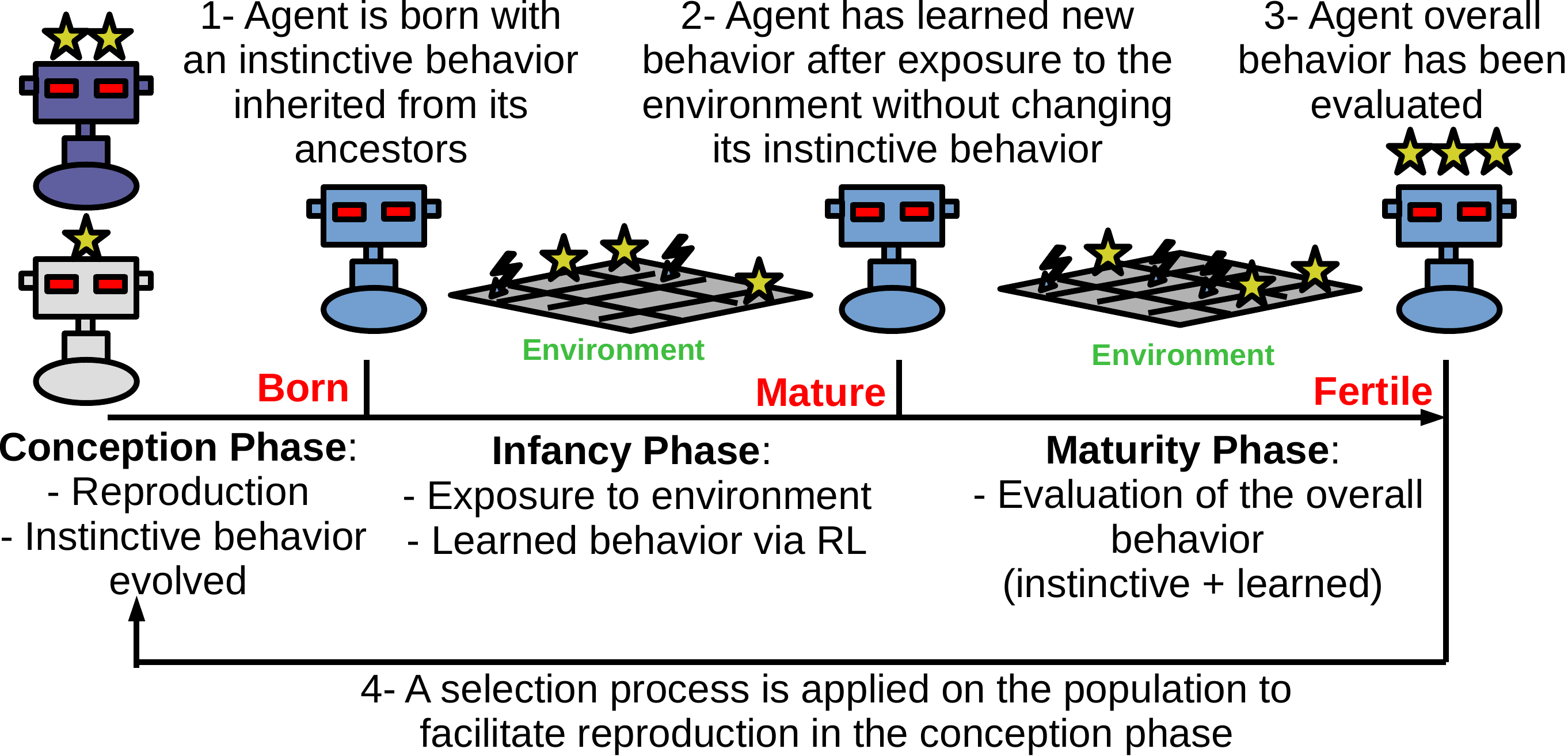}
        \caption{The Evo-RL scheme showing the agent life-cycle, highlighting its different phases and states (born, mature and fertile).}
        \label{fig:lifeCycle}
    \end{figure}%

The outer loop of Evo-RL is an evolutionary algorithm. Similar to any EA, Evo-RL starts by initializing a set of individuals (agents), thus, producing a population of agents. In the first iteration, each agent has a randomly initialized behaviour. This behaviour has a dual representation, one observable \emph{phenotype} and its corresponding \emph{genotype}. The genotype encodes the phenotype and contains all the information necessary to build the phenotype. For example, the phenotype can be an Artificial Neural Network (ANN) representing the agent behaviour, while the genotype can be a set of binary numbers that when decoded produce the ANN of that agent's behaviour. The genotype is necessary for reproducing newly born agents in a later stage.

After this initialization, each agent in the population is considered in the born state, and has an instinctive behaviour. As shown in Figure \ref{fig:lifeCycle}, after birth, the agent starts to be exposed to the environment. In this \textit{infancy phase}, reinforcement learning is executed. However, the agent cannot overwrite its instinctive behaviour, thus, if the agent is in a state where the instinctive behaviour has already defined an action to execute, the agent executes this action and no learning is done with respect to this state. On the other hand, if the agent is in a state where the instinctive behaviour does not define what should be done, then the agent proceeds with its learning algorithm normally. After the infancy phase ends, due to resource or time constraints for example, the agent reaches the mature stage and is now ready to be evaluated. In the \textit{maturity phase}, the agent overall behaviour is evaluated with respect to pre-set objectives. A score that measures its performance is then calculated.

Based on this score, a selection process of parents is conducted, taking into consideration all agents in the population. This selection process must ensure diversity to avoid getting stuck on local maxima. Based on selection procedure, we reach now the \textit{conception phase}, where different reproduction operators are executed on the genotype, for instance by combining instinctive behaviour genes from different parents. Furthermore, to allow plasticity of learned behaviour, in the conception phase, parents also exchange their learned behaviour. For example, if the learned behaviour is an artificial neural network (ANN), then the weights of the network of the parents is averaged, thus, allowing the next generation to gain from past learned experiences of their parents. However, this process does not intervene with the evolutionary process, as the exchange of genes is done only on the instinctive behaviour. Furthermore, the instinctive behaviour overwrites any learned behaviour. Hence, if the instinctive behaviour choose to handle a particular state, no learning is done on that state.

Finally, after conception, a set of newly born agents with evolved instinctive behaviours are now ready for a new cycle where they are exposed to the environment. All these steps can be repeated until a stopping criterion is reached, such as exhausting a pre-defined computational budget.

To summarize, the overall approach is an evolutionary computation approach. However, the novelty can be highlighted in the following design axioms:
\begin{enumerate}
\item The choice of which part of the overall behaviour is instinctive, and which is not, is decided by the evolutionary process. In other words, the line between what is instinctive (fully evolvable) and learnable, is evolved. This is achieved by not allowing the learning process to overwrite the instinctive behaviour. Evolution dictates which region of states it operates on. 
\item The overall fitness of an agent considers both behaviors, i.e., instinctive plus learnable. This is facilitated by conducting the evaluation of the behaviour after the learning process is conducted.
\item In the conception phase, only the instinctive behaviour is evolved, but the learned behaviour is transferred to the off-springs to allow plasticity in the learned behaviour as long as the instinctive behaviour allows it. In case of conflict, the instinctive behaviour overwrites any learnable behaviour.
\end{enumerate}

\section{Related Work}

Generally, we propose an algorithm that encapsulates reinforcement learning within an evolutionary algorithm cycle. The importance of combining evolutionary computation with learning in general was highlighted in \cite{hinton1987learning}. In this paper, authors argued that adopting a genetic algorithm combined with a local and myopic optimizer such as hillclimbing will lead to a better search algorithm than either of these algorithms alone.

In recent years, researchers proposed different approaches to combine evolutionary computation with reinforcement learning. Notably, one approach is realized by combining deep neuroevolution and deep reinforcement learning \cite{DBLP:journals/corr/abs-1810-01222}. In this work, authors use a simple cross-entropy method (CEM) in combination of an off-policy deep reinforcement learning.

In \cite{DBLP:journals/corr/abs-1906-09807}, authors propose a novel algorithm called Proximal Distilled Evolutionary Reinforcement Learning (PDERL). In their approach, they adopt a hierarchical integration between evolution and learning.

Furthermore, authors in \cite{khadka2019collaborative} combined EA and RL by introducing a Collaborative Evolutionary Reinforcement Learning (CERL) algorithm. In CERL, different learners with different time horizons explore the solution space while committing to the task they are solving.
\section{Evo-RL Implementation}
The approach presented in the previous section can be implemented in various ways. For example, we do not dictate a particular way for representing the instinctive behaviour or the learnable behaviour. Furthermore, this approach works with any reinforcement learning algorithm; likewise, any evolutionary algorithm can be adopted.

\begin{figure}[h]
    \centering
    \includegraphics[width = \linewidth]{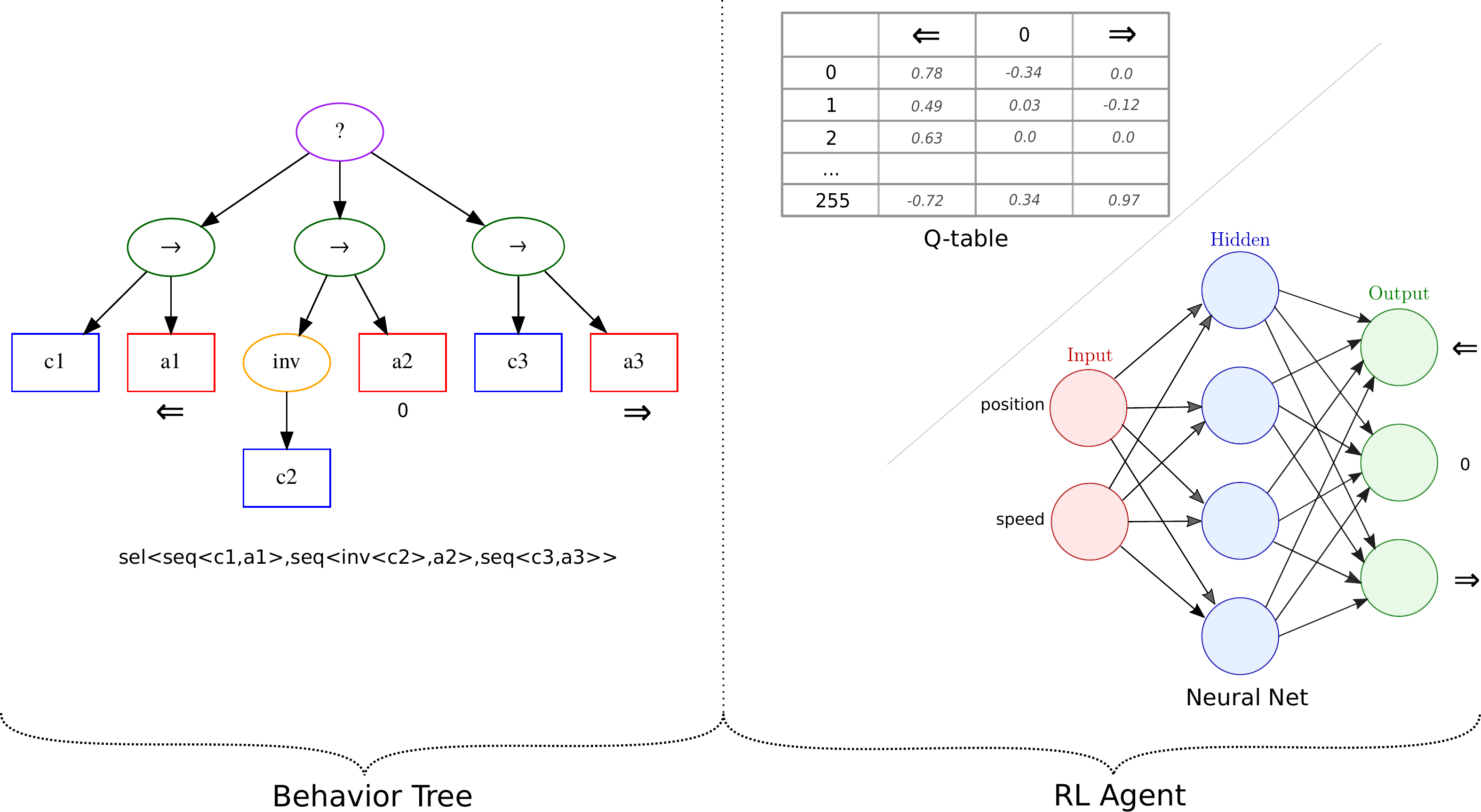}
    \caption{Example representation of an agent behaviour. Left: Instinctive behaviour in the form of BT. Right: Two possible learned behaviour representations, a tabular form and an ANN. Note that this is a simplified version of an ANN.
    }
    \label{fig:implementation:individual}
\end{figure}

However, for evaluation purposes, we have implemented the algorithm as follows. Firstly, we used the EA in the form of Genetic Programming (GP). However, as for what concerns the representation of the instinctive (evolved) behaviour, we adopted \textit{behaviour trees} (BTs). These fit well with GP and, unlike ANN, are much easier to interpret. As for the learned behaviour, we adopted two possibilities, one tabular representation used when testing our approach with Q-learning, and another ANN representation when testing our approach with Proximal Policy Optimization (PPO) and Deep Q-Network (DQN) algorithm. Figure \ref{fig:implementation:individual} shows an example of the overall behaviour of an agent.

\subsection{Behavior Trees}
\label{lab:bg:BT}
A BT is a formal, graphical modelling language that was introduced in the gaming industry in order to simplify the programming of bots.
The behavior of these bots is described in the form of a graph that represents a tree that controls the flow of decision making. This tree is a depth-first acyclic directed graph. The top most node is called the \emph{root node}, which is the only node that does not have a parent. 

During execution, the BT gets ticked starting at the \emph{root node} every time. It then traverses the tree in a depth-first order, ticking each node.
Depending on the signal generated by the children of the currently ticked node, a certain signal is generated that determines the next node to tick.
Executing the BT again will tick all the nodes again, i.e. there is no memory in between different ticks. A signal can take one the following values:

\begin{itemize}
  \item Success: Informs the parent that the command was a success.
  \item Failure: Informs the parent that the command was a failure.
  \item Running: Informs the parent that the command is still running.
\end{itemize}

The nodes of a BT can either be \emph{leaf nodes} or \emph{composite nodes}.

\begin{figure}[!ht]
    \centering
    \begin{subfigure}[t]{0.12\textwidth}
        % {\includegraphics[scale=0.65]{figures/BT/selector.png}}
        \includegraphics[width=1\textwidth]{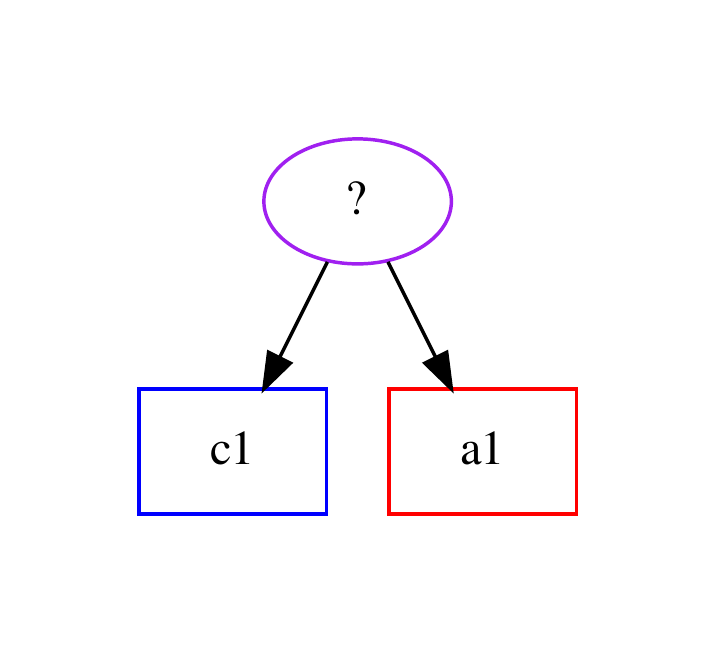}
        \caption{Selector}
        \label{fig:bg:composite_nodes:Selector}
    \end{subfigure}%
    % \\
    \begin{subfigure}[t]{0.12\textwidth}
        % {\includegraphics[scale=0.65]{figures/BT/sequence.png}}
        \includegraphics[width=1\textwidth]{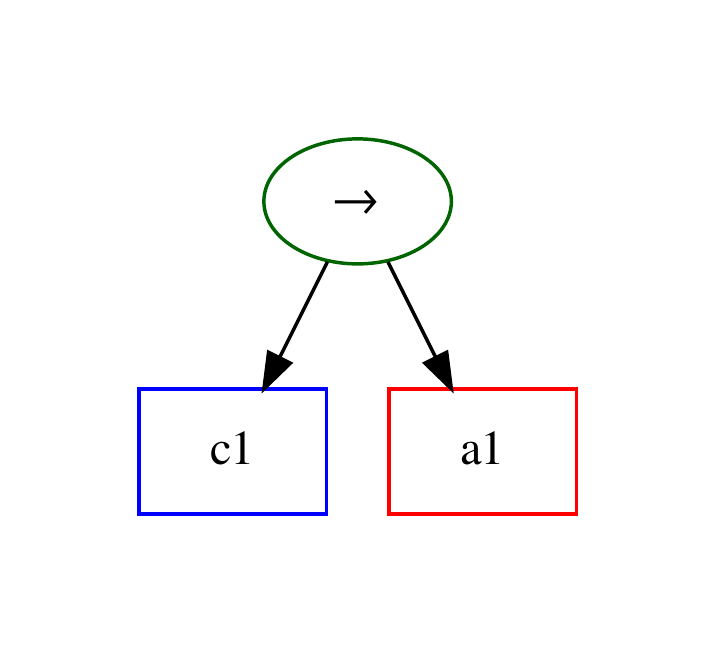}
        \caption{Sequence}
        \label{fig:bg:composite_nodes:Sequence}
    \end{subfigure}%
   \begin{subfigure}[t]{0.12\textwidth}
        % {\includegraphics[scale=0.65]{figures/BT/decorator.png}}
        \includegraphics[width=1\textwidth]{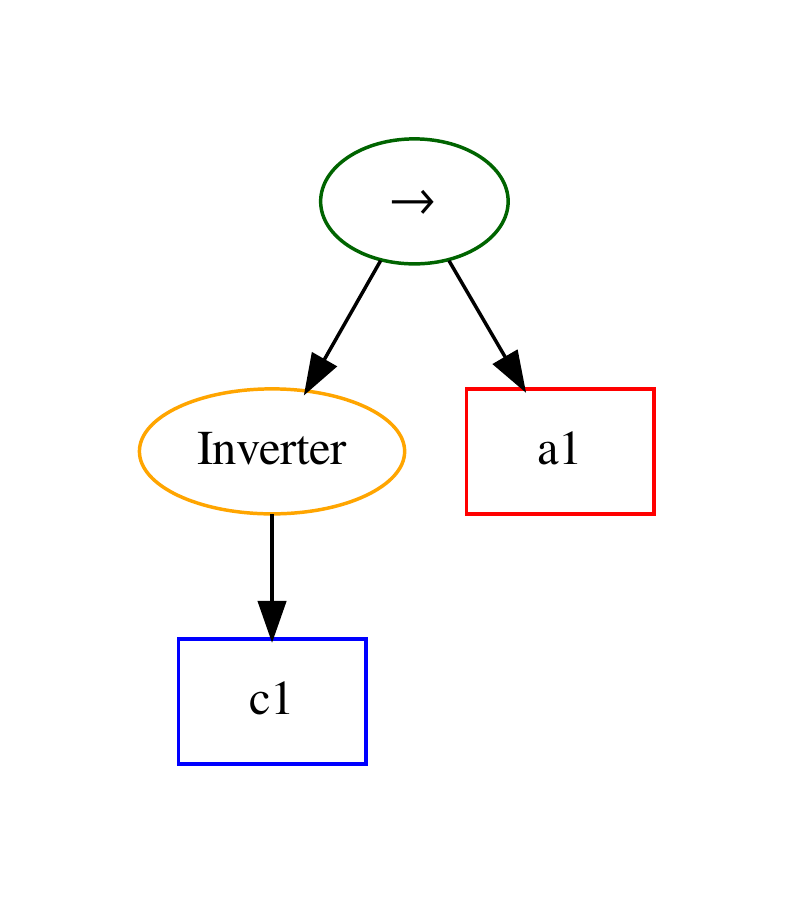}
        \caption{Decorator}
         \label{fig:bg:composite_nodes:Decorator}
    \end{subfigure}%
    \begin{subfigure}[t]{0.12\textwidth}
        \includegraphics[width=1\textwidth]{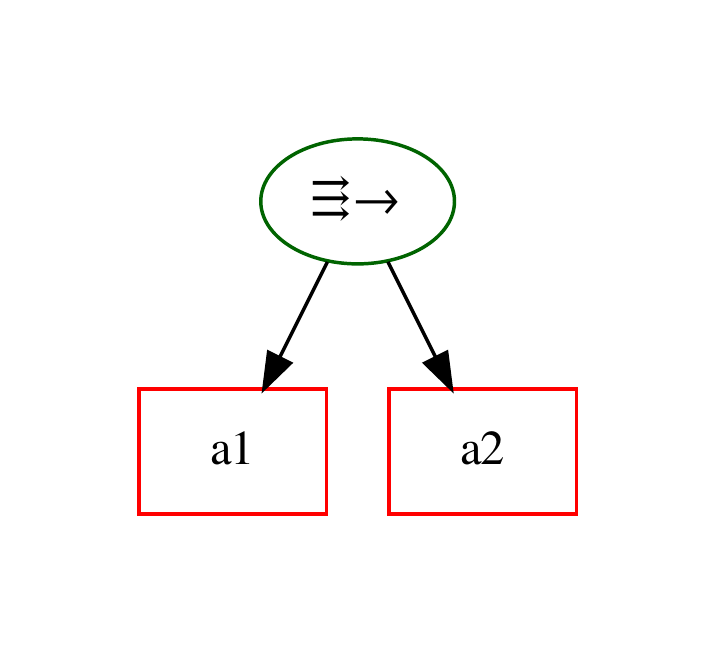}
        \caption{Decorator}
        \label{fig:bg:composite_nodes:Parallel}
    \end{subfigure}%
    \caption{Composite nodes in Behavior Trees}
    \label{fig:bg:composite_nodes}
\end{figure}

Composite nodes can have one or more children and are parents to other composite nodes or to a leaf node. They can be any of the following types: \emph{Selector}, \emph{Sequence}, \emph{Decorator}, or \emph{Parallel}. A selector node visits all of its children in order, from left to right. If any of its children returns the status \emph{Success}, \emph{Success} is sent upwards in the tree. If all children were ticked and none of them returned \emph{Success}, the signal \emph{Failure} is set upwards in the tree. This node is also sometimes called a fallback node. It is represented as an oval with a question mark~(\Cref{fig:bg:composite_nodes:Selector}). A sequence node visits all of its children in order, from left to right. If and only if all children return the status \emph{Success}, \emph{Success} is also propagated upwards in the tree. If any child returns \emph{Failure}, the status \emph{Failure} is immediately sent upwards in the tree. It is represented as an oval with a right pointing arrow~(\Cref{fig:bg:composite_nodes:Sequence}). The decorator nodes only have one child and their purpose is to modify the signal based on its type~(\Cref{fig:bg:composite_nodes:Decorator}). For example, an \textit{invert} node, negates the signal generated by its child. Thus \emph{Success} will become \emph{Failure}, and vice versa. A \textit{Repeater} repeats ticking its child $x$ times. \textit{Repeat Until Fail} ticks its child in a loop until it returns \emph{Failure}. On the other hand, in parallel nodes, all of its children get ticked at the same time~(\Cref{fig:bg:composite_nodes:Parallel}). This allows multiple children to enter the running state. The signal that gets propagated upwards depends on whether a \emph{Parallel-Selector} or a \emph{Parallel-Sequence} is used. \textit{Parallel-Selector} return \emph{Success} if any of its children returned \emph{Success}, otherwise returns \emph{Failure}. \textit{Parallel-Sequence} returns \emph{Success} if all of its children returned \emph{Success}, otherwise returns \emph{Failure}.
\vspace{-1em}
%\vspace{-1em}
Leaf nodes of BTs are input-output nodes that interact with an environment. A leaf node can either be a \emph{condition node} or an \emph{action node}. 

\emph{Condition nodes} check whether a condition is satisfied, and return the status \emph{Success} in that particular case. A \emph{condition node} only passively observes the environment and never returns the signal \emph{Running}.
A condition might consist of several sub-conditions or sensor values coming from multiple independent or dependent features.

\emph{Action nodes} execute an actual action in an environment. Executing an action in an environment changes the state of the environment and therefore impacts the subsequent nodes within the same tick. An action node returns \emph{Success} if the action is completed, \emph{Failure} if it couldn't get completed, and  \emph{Running} if the action did not finish immediately and is still running. When combining \emph{condition nodes} and \emph{action nodes}, arbitrary instinctive behaviors can be described with a BT~\cite{paper:instinct_driven}.
    \newpage
    \section{Experimental Evaluation}
        We designed our experimental evaluation to test the following statements: (1) The performance of reinforcement learning algorithms is enhanced when embedded in the Evo-RL approach for environments with rewardless states, with the same fixed computational budget. (2) The performance of Evo-RL is better than the evolutionary algorithm part alone (i.e., Evo-RL without the reinforcement learning). In other words, we want to show that instinctive behaviour plus learnable behaviour (Evo-RL) outperforms adopting only instinctive behaviour (EA-only) or only learnable behaviour (RL-only). (3) As the rewardless states increase in an environment, the ratio of instinctive behaviour executed, compared to the learnable one, increases as well. This shows that the instinctive behaviour is necessary to handle more efficiently the rewardless states.
    \begin{figure}[!ht]
    \centering
    \includegraphics[width= 0.4\textwidth]{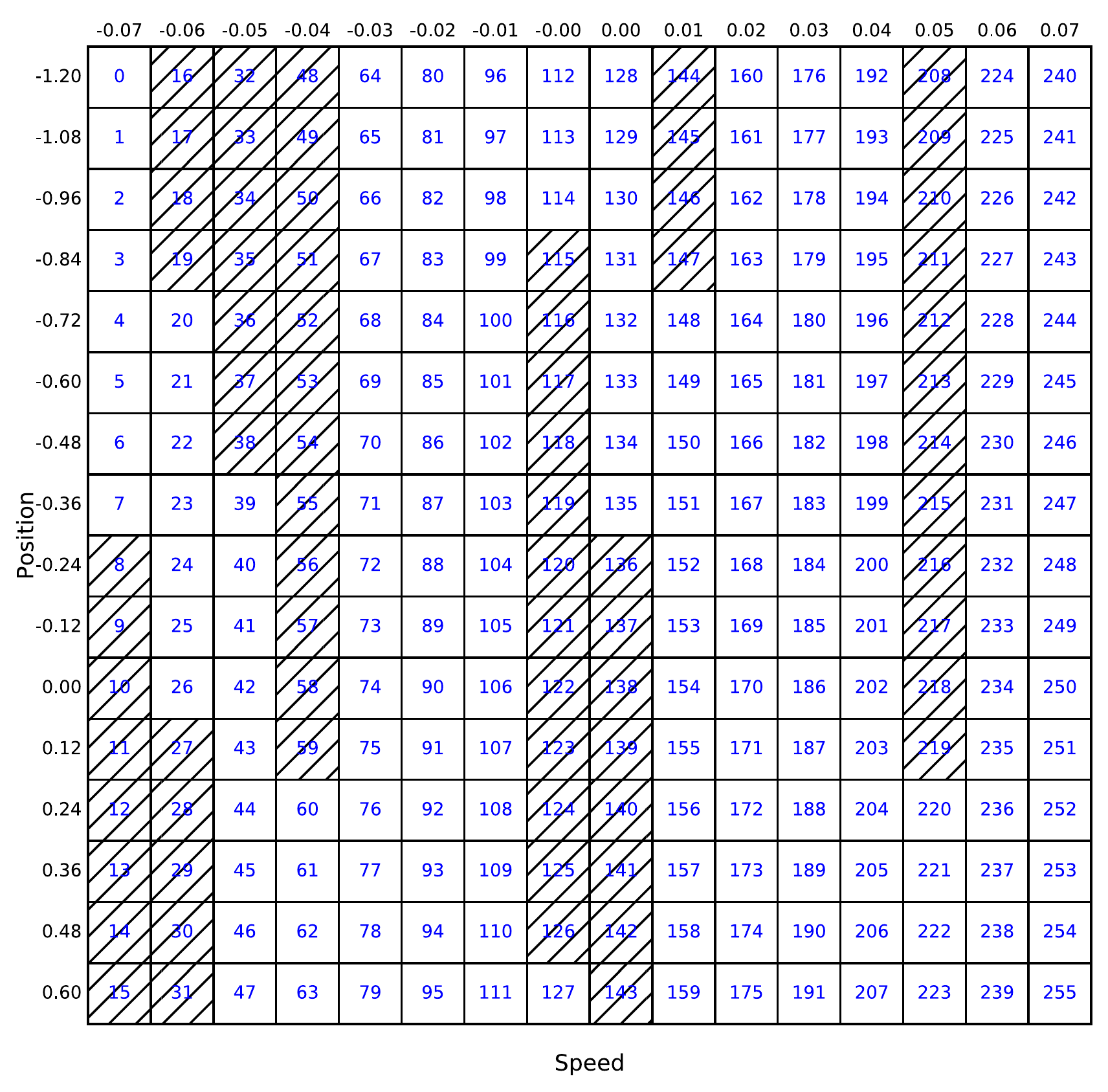}
    \caption{Example of marking 30\% of all states as rewardless for the \emph{MountainCar} problem (2D state space). The overall state space is binned into 256 bins. Hashed states denote a rewardless state.}
    \label{fig:implementation:rewardless_states}
    \end{figure}
    \begin{figure*}[!ht]
    \centering
    \begin{subfigure}[t]{0.33\textwidth}
        \centering
        \includegraphics[width=1\textwidth]{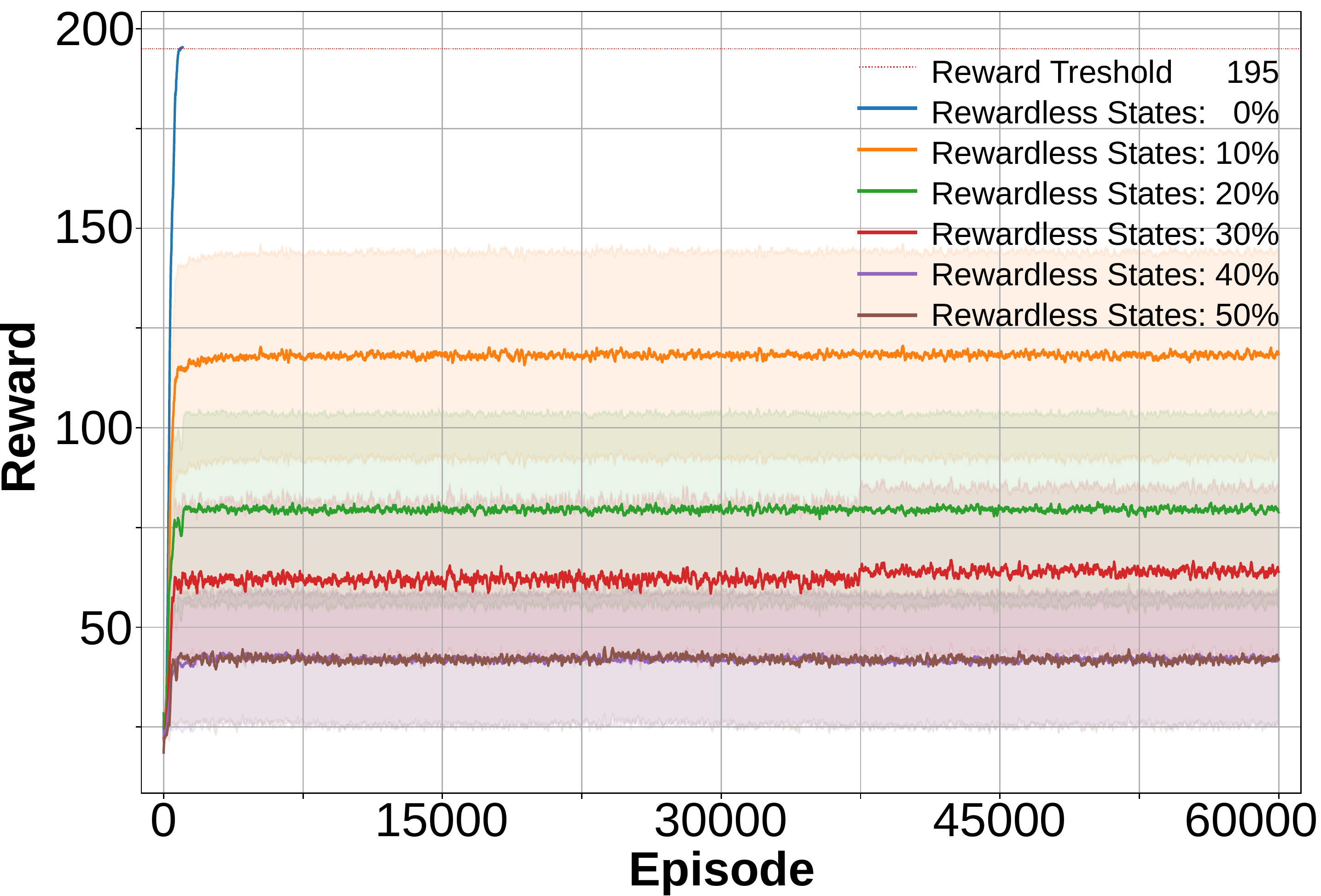}
        \caption{Q-learning (CartPole)}
    \end{subfigure}%
    \begin{subfigure}[t]{0.33\textwidth}
        \centering
        \includegraphics[width=1\textwidth]{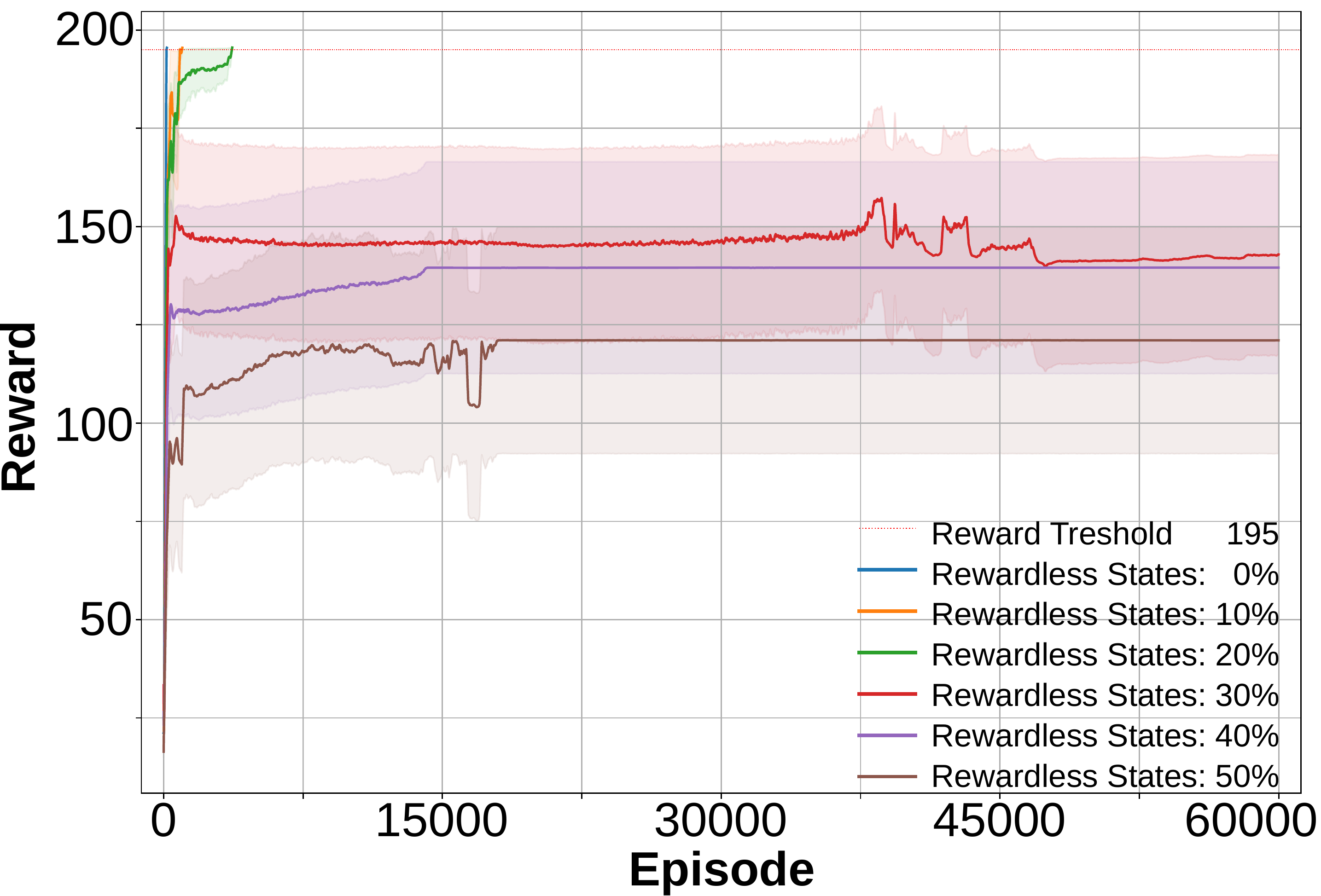}
        \caption{DQN (CartPole)}
    \end{subfigure}%
        \begin{subfigure}[t]{0.33\textwidth}
        \centering
        \includegraphics[width=1\textwidth]{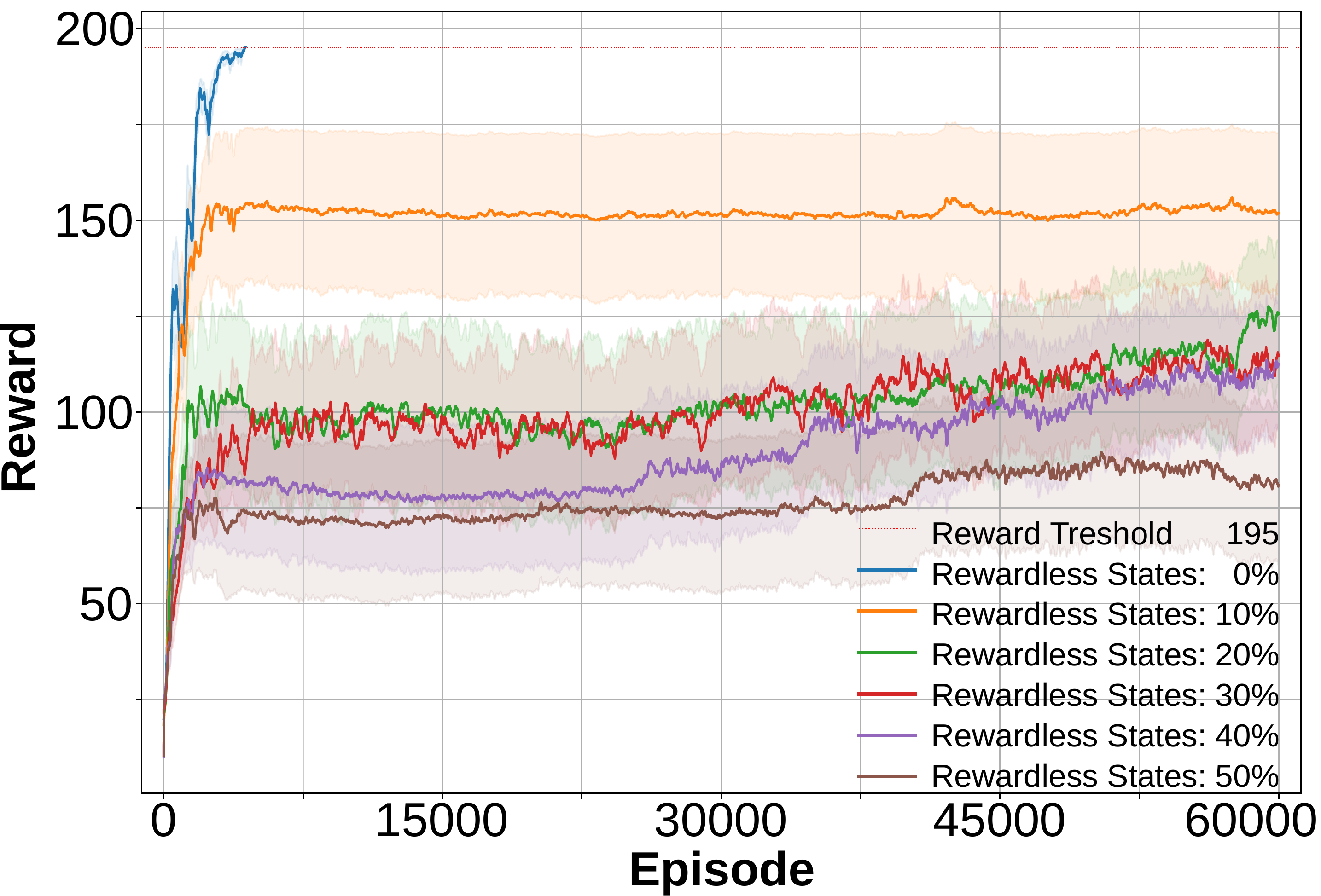}
        \caption{PPO (CartPole)}
    \end{subfigure}%
    \\
    \begin{subfigure}[t]{0.33\textwidth}
        \centering
        \includegraphics[width=1\textwidth]{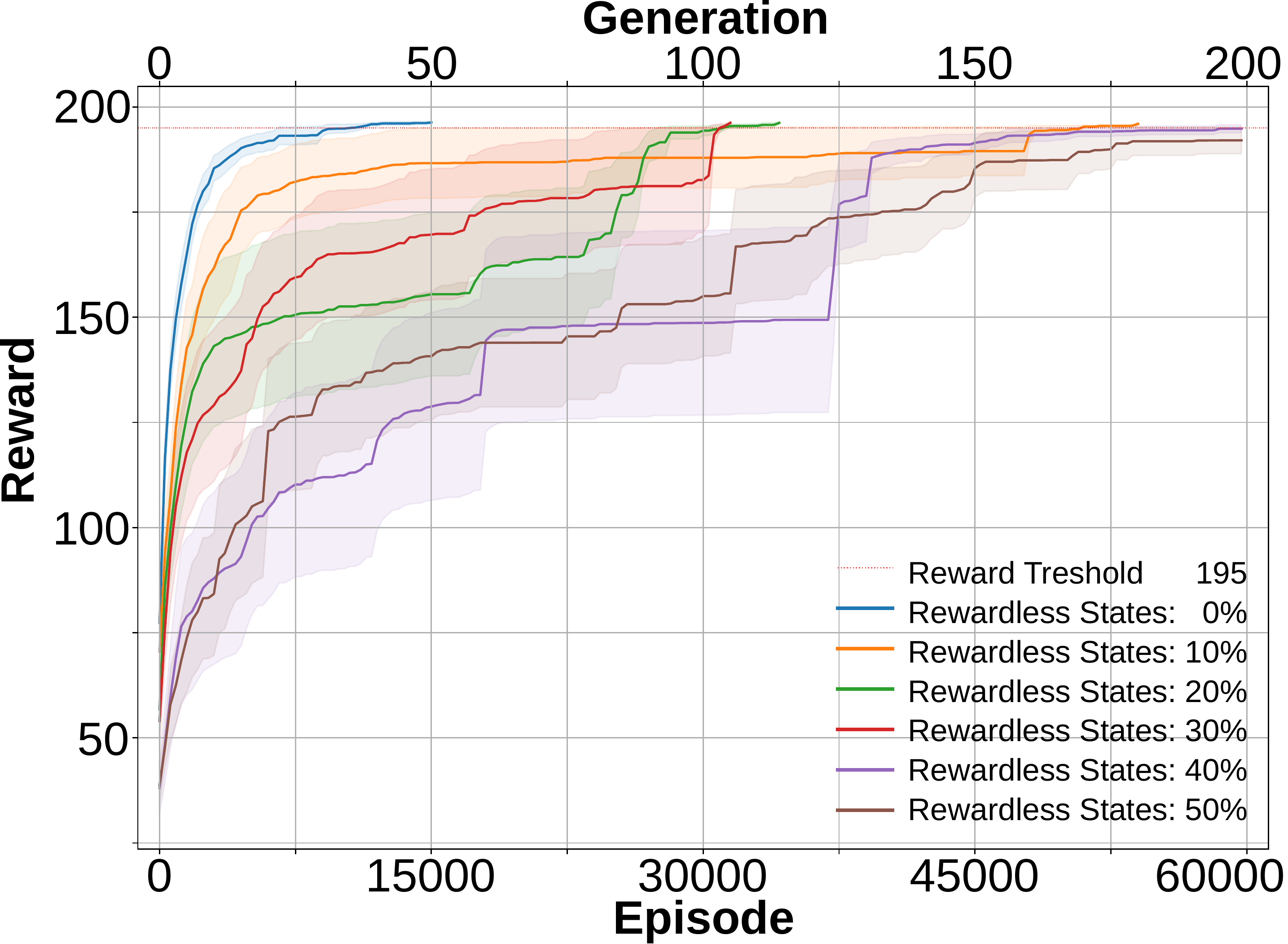}
        \caption{e-Q learning (CartPole)}
    \end{subfigure}
         \begin{subfigure}[t]{0.33\textwidth}
        \centering
        \includegraphics[width=1\textwidth]{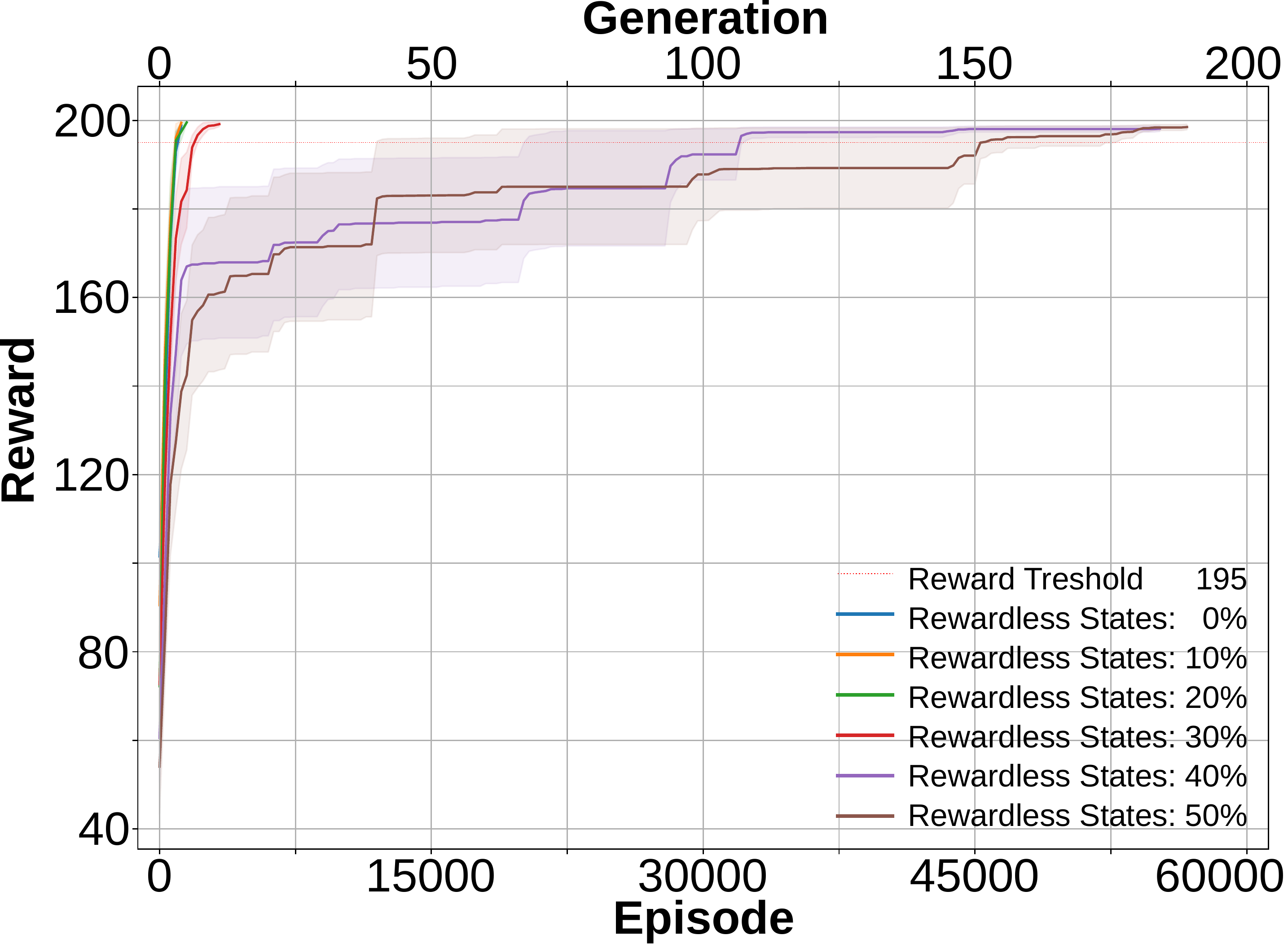}
        \caption{e-DQN (CartPole)}
    \end{subfigure}
    \begin{subfigure}[t]{0.33\textwidth}
        \centering
        \includegraphics[width=1\textwidth]{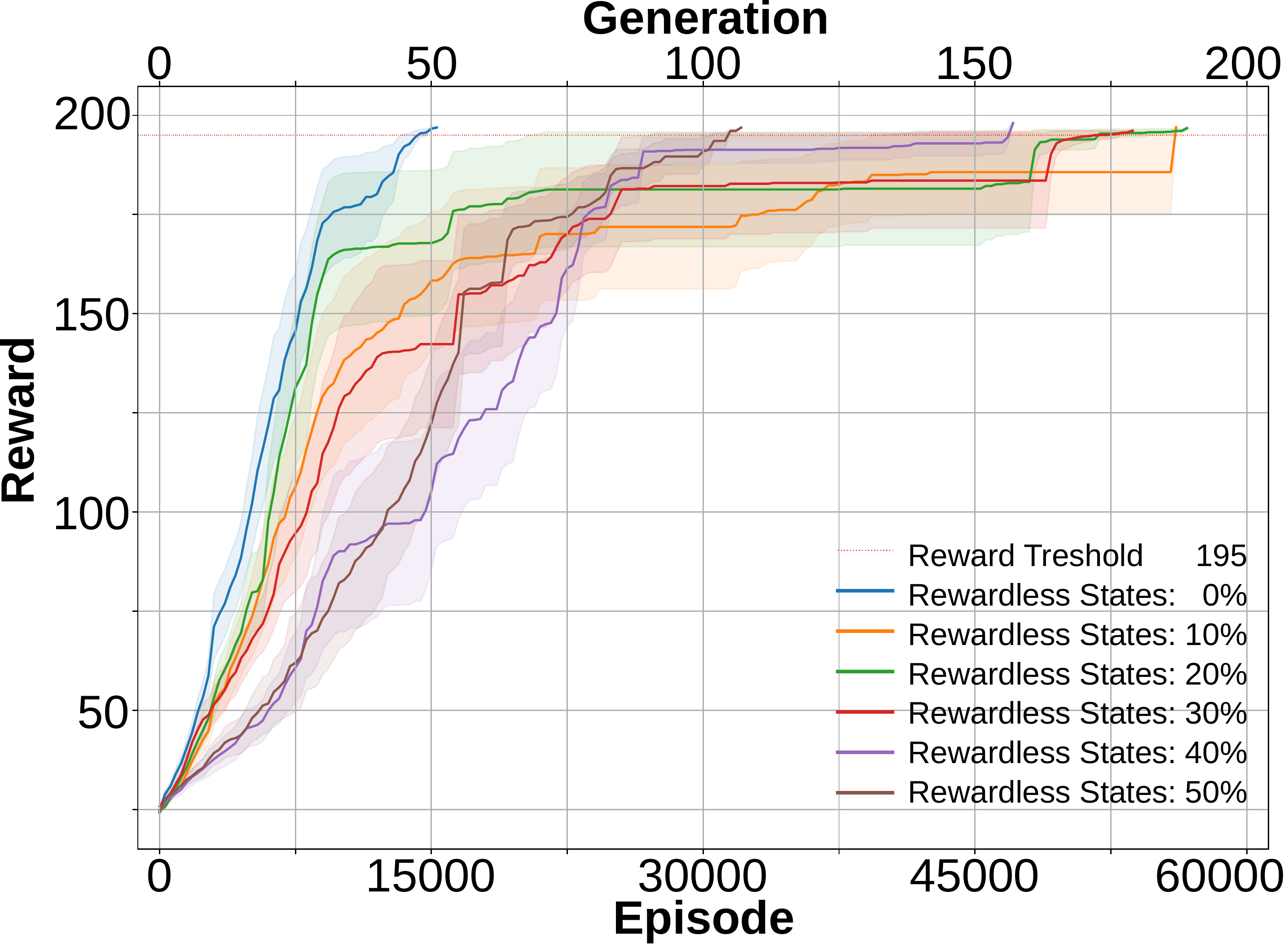}
        \caption{e-PPO (CartPole)}
    \end{subfigure}
    \begin{subfigure}[t]{0.33\textwidth}
        \centering
        \includegraphics[width=1\textwidth]{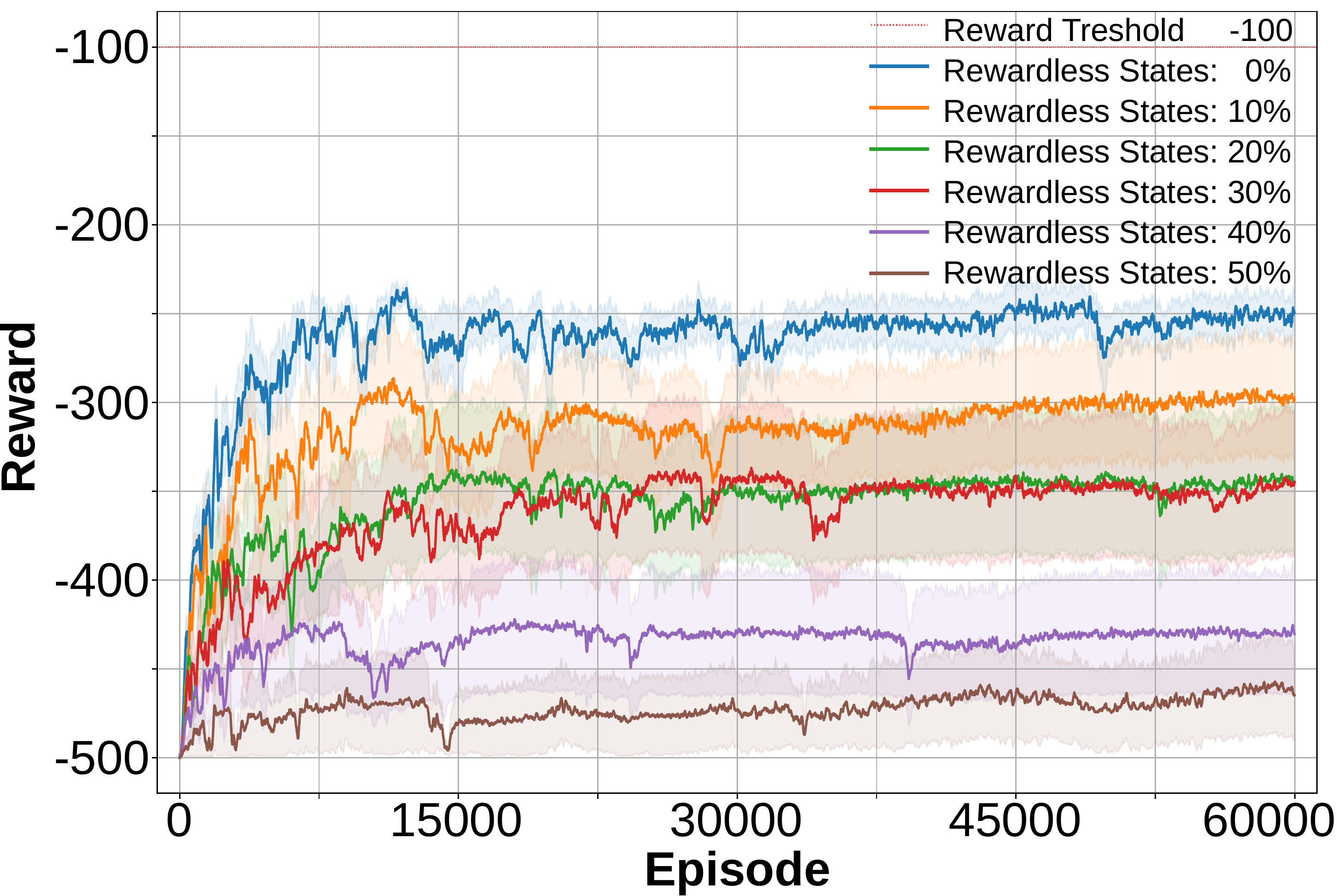}
        \caption{Q learning (Acrobot)}
    \end{subfigure}%
    \centering
    \begin{subfigure}[t]{0.33\textwidth}
        \centering
        \includegraphics[width=1\textwidth]{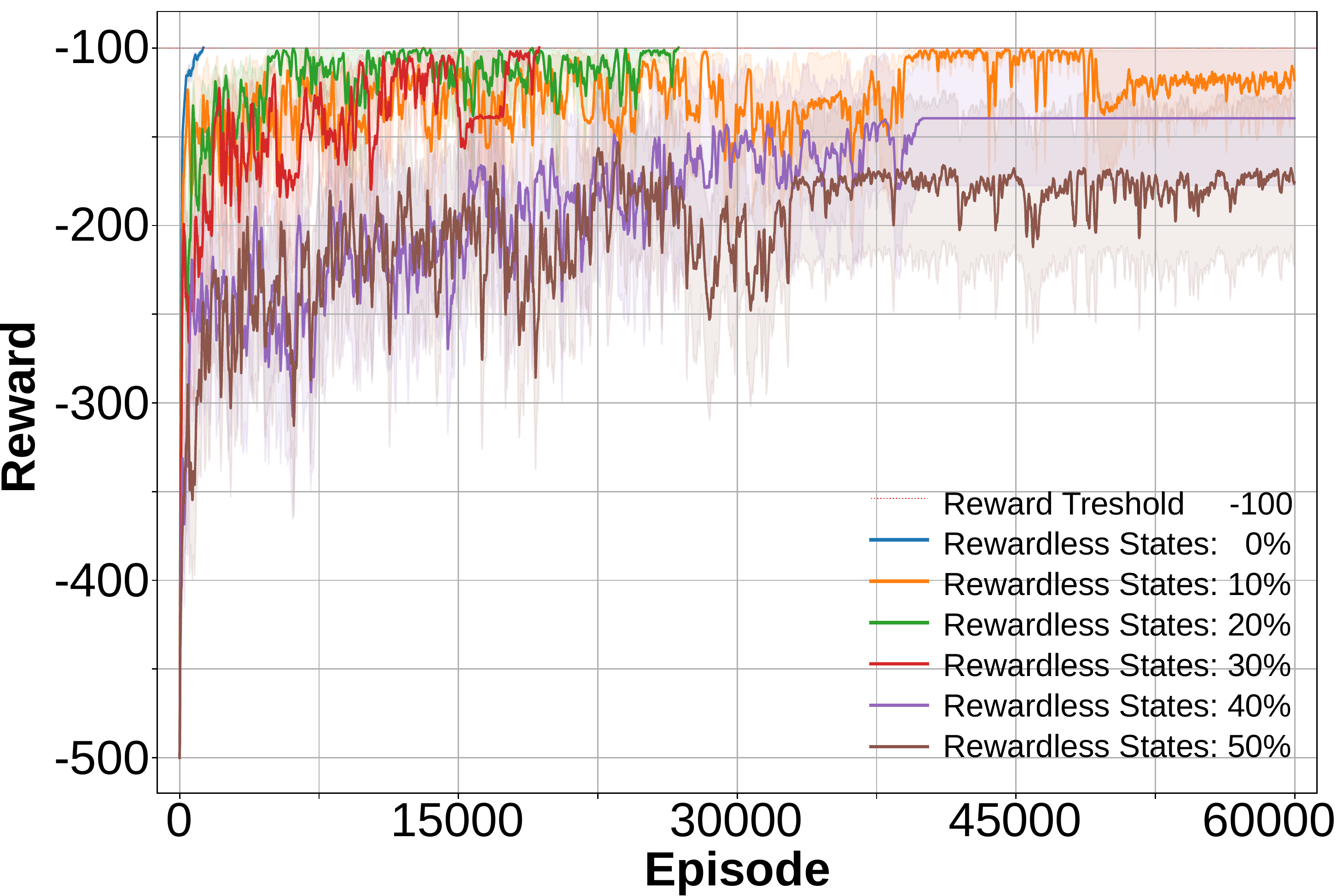}
        \caption{DQN (Acrobot)}
    \end{subfigure}%
        \begin{subfigure}[t]{0.33\textwidth}
        \centering
        \includegraphics[width=1\textwidth]{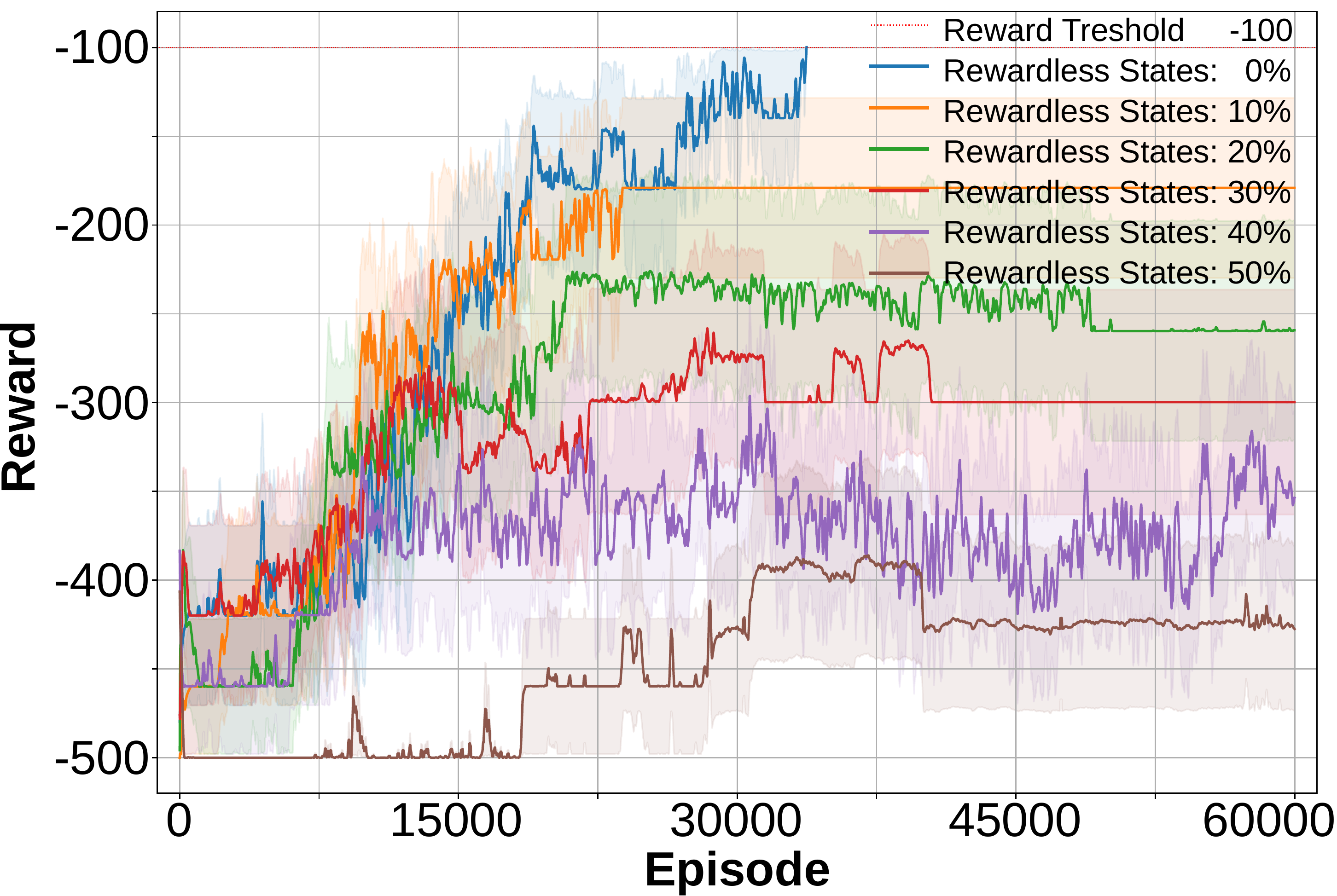}
        \caption{PPO (Acrobot)}
    \end{subfigure}%
    \\
    \begin{subfigure}[t]{0.33\textwidth}
        \centering
        \includegraphics[width=1\textwidth]{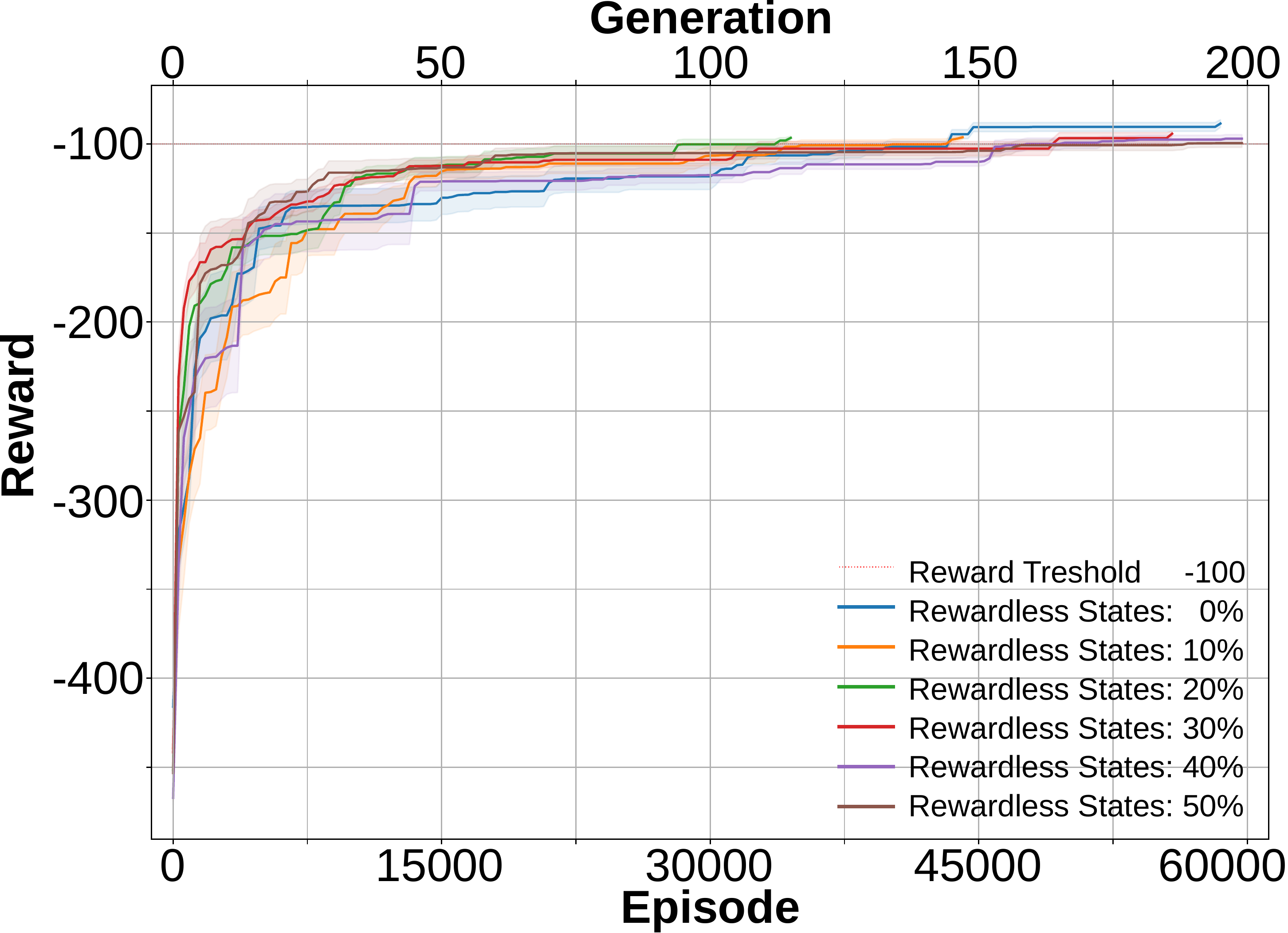}
        \caption{e-Q learning (Acrobot)}
    \end{subfigure}
    \begin{subfigure}[t]{0.33\textwidth}
        \centering
        \includegraphics[width=1\textwidth]{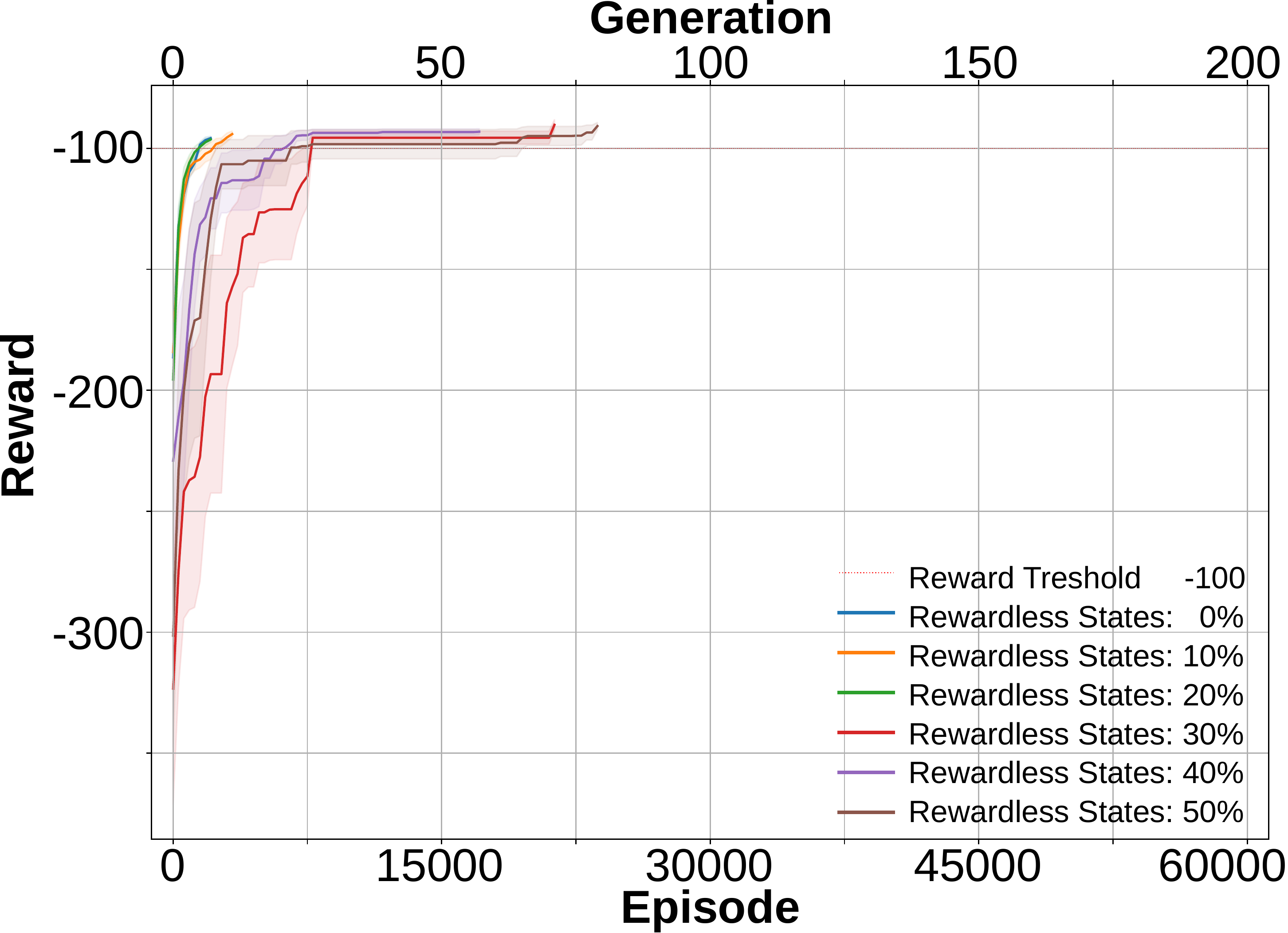}
        \caption{e-DQN (Acrobot)}
    \end{subfigure}
     \begin{subfigure}[t]{0.33\textwidth}
        \centering
        \includegraphics[width=1\textwidth]{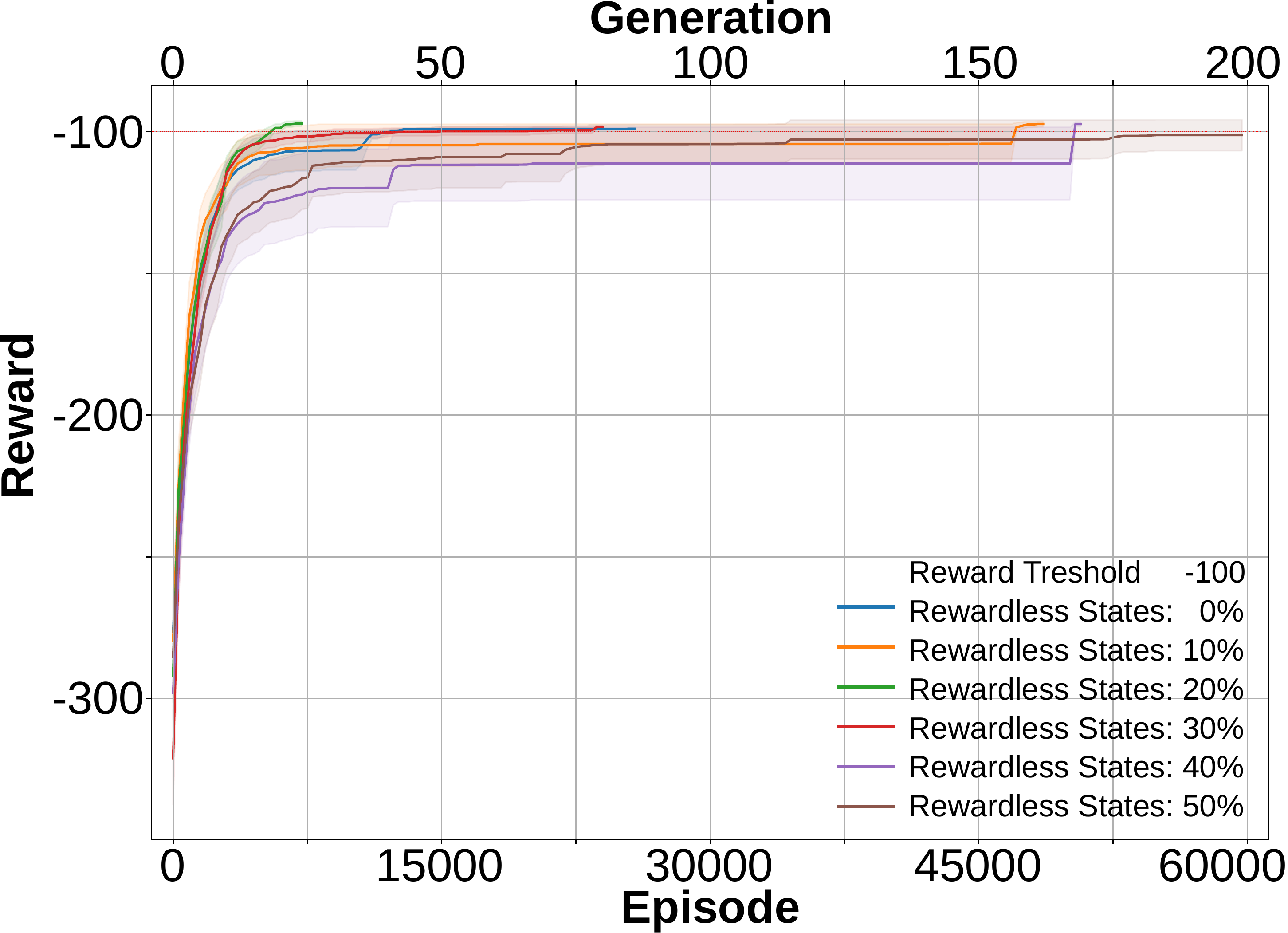}
        \caption{e-PPO (Acrobot)}
    \end{subfigure}
    \caption{Results of the CartPole and Acrobot problems (mean and SEM over 10 trials). Each algorithm is tested on different percentages of rewardless states: 0\%, 10\%, 20\%, 30\%, 40\% and 50\%. Each case is presented by a different color. Generally, as the percentage of rewardless states increases, the performance of the algorithm deteriorates. The eRL version for all algorithms outperforms the corresponding (RL-only) stand-alone version.}
    \label{fig:cartpole}
    \end{figure*}
     \begin{figure*}[!ht]
    \centering
    \begin{subfigure}[t]{0.33\textwidth}
        \centering
        \includegraphics[width=1\textwidth]{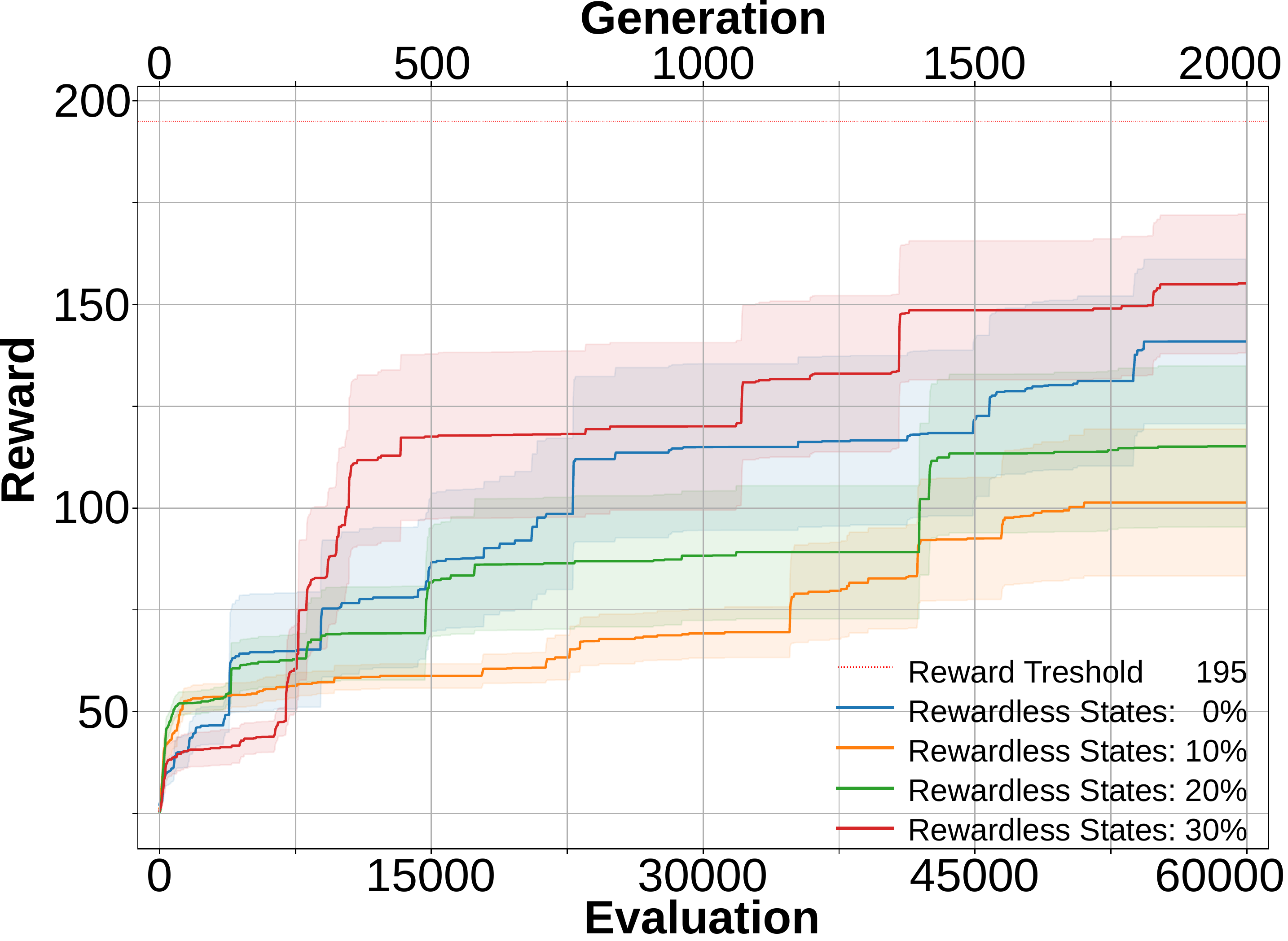}
        \caption{Cart Pole Problem}
    \end{subfigure}%
    \begin{subfigure}[t]{0.33\textwidth}
        \centering
        \includegraphics[width=1\textwidth]{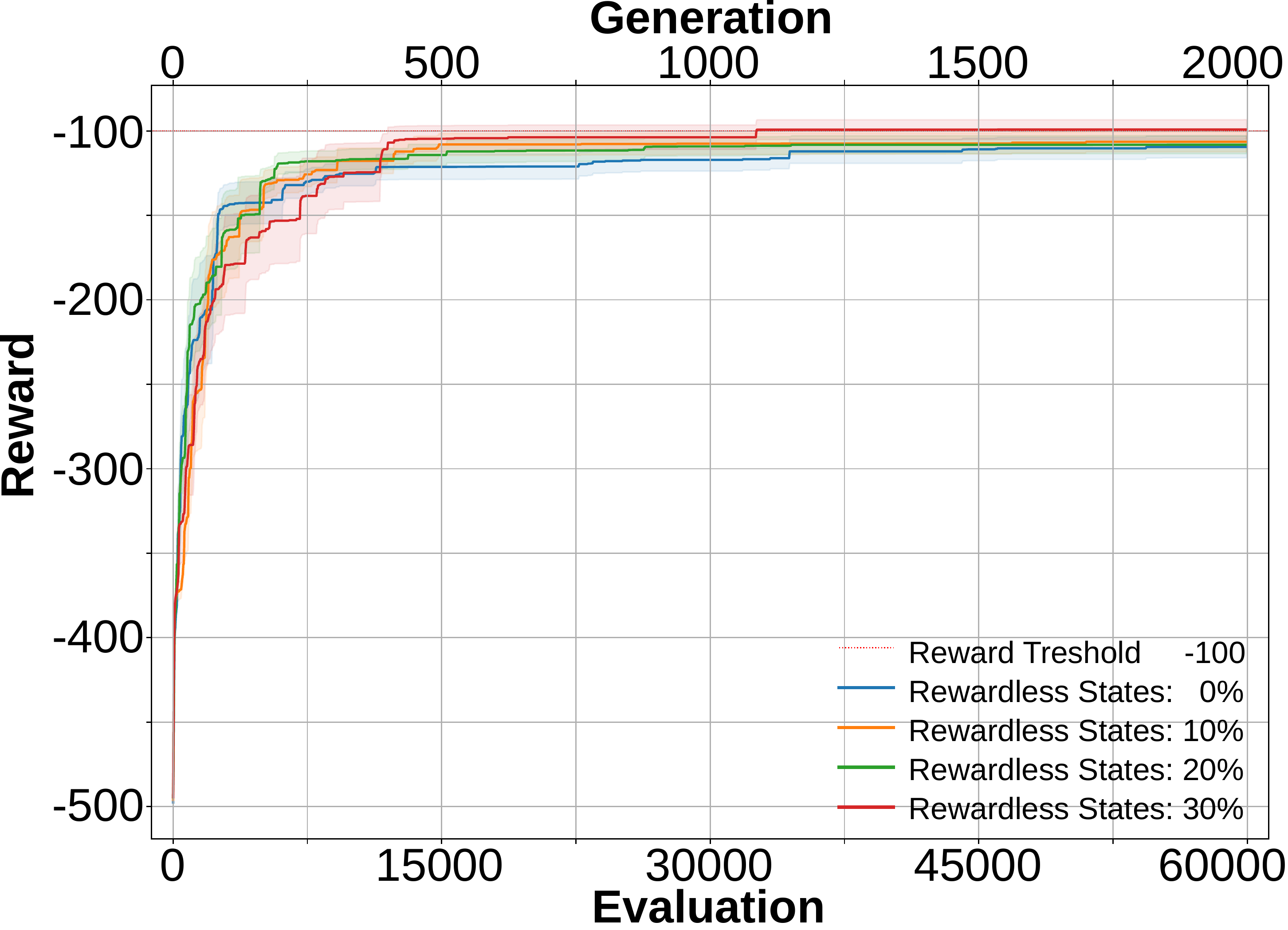}
        \caption{Acrobot Problem}
    \end{subfigure}
        \begin{subfigure}[t]{0.33\textwidth}
        \centering
        \includegraphics[width=1\textwidth]{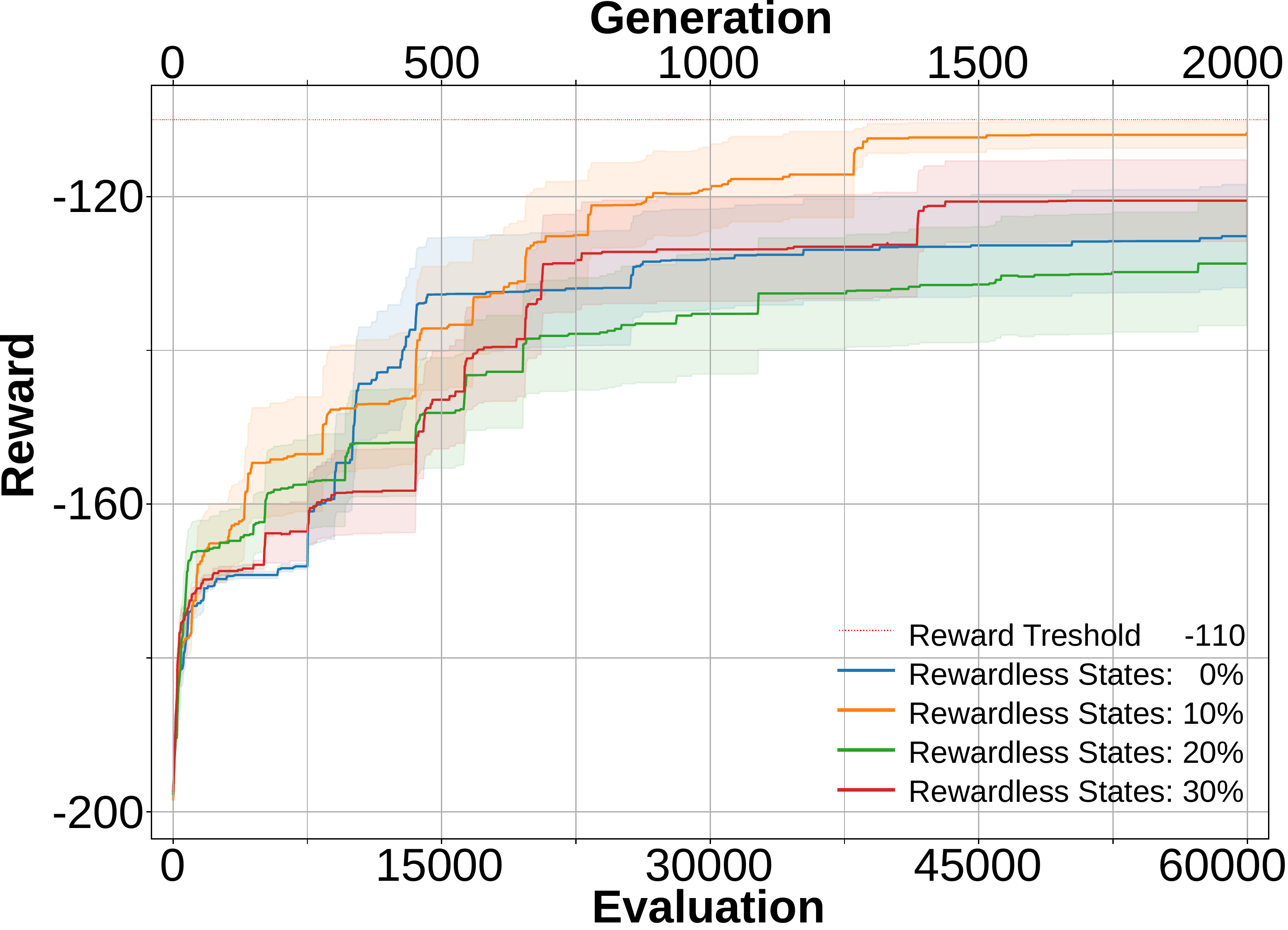}
        \caption{MountainCar Problem}
    \end{subfigure}%
    \caption{Results of the CartPole, Acrobot and MountainCar problems in case of using EA-only for different percentages of rewardless states: 0\%, 10\%, 20\%, 30\%, 40\% and 50\% (mean and SEM over 10 trials).}
    \label{fig:EA_only}
 \end{figure*}

     \begin{figure}[!ht]
    \centering
    \includegraphics[width=0.9\linewidth]{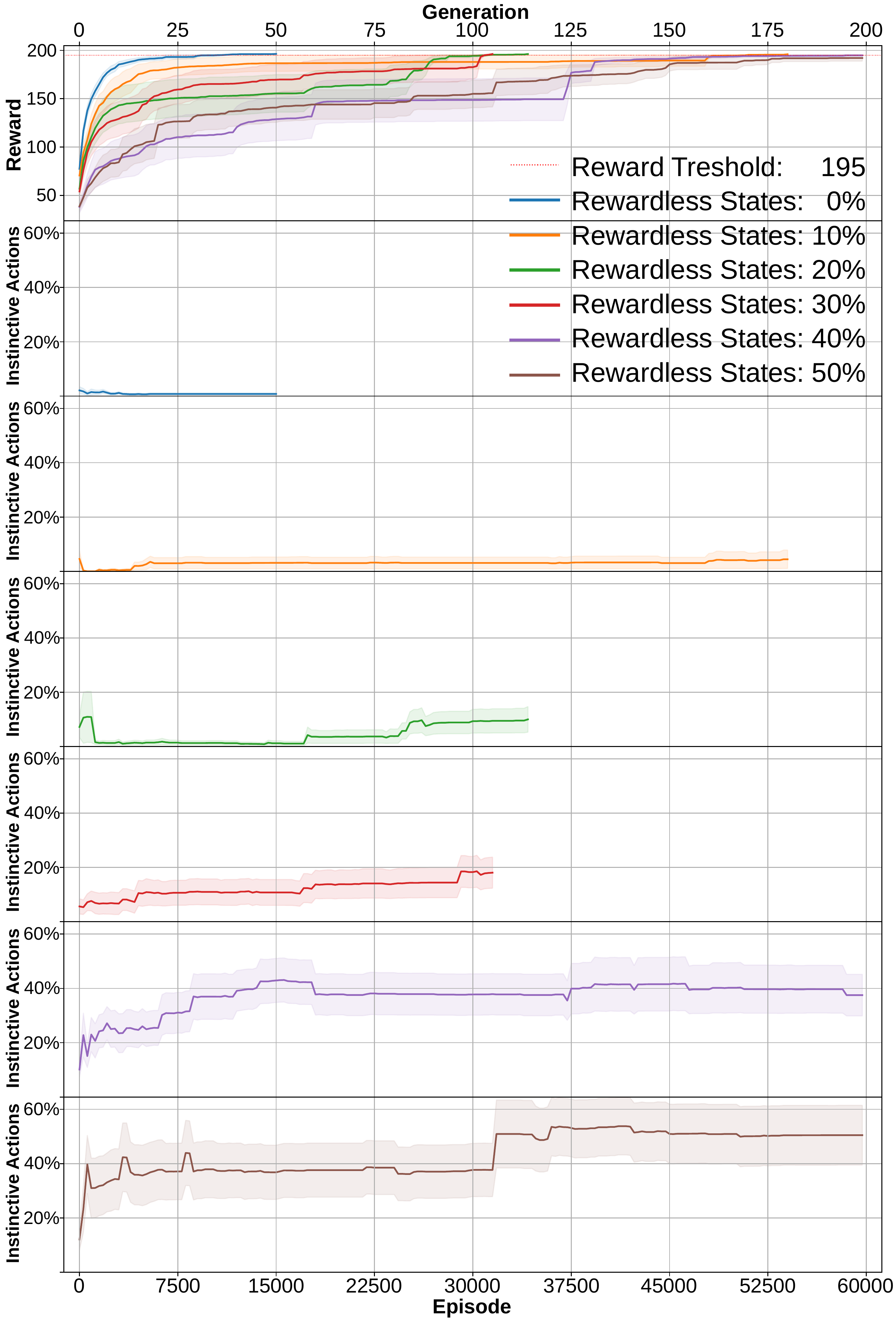}
    \caption{Top: convergence plot of e-Q learning with different on the CartPole problem with rewardless states: 0\%, 10\%, 20\%, 30\%, 40\% and 50\%. The remaining plots present the ratio of actions executed by the instinctive behaviour when solving the CartPole problem for environments with different rewardless states 0\%, 10\%, 20\%, 30\%, 40\% and 50\%. As the percentage of rewardless states increases, the behaviour relies more on the instinctive behaviour.}
    \label{fig:evaluation:CartPole:EA-PPO:multi}
    \end{figure}

    In order to facilitate testing on environments with a rewardless state, we modified three OpenAI gym control problems\footnote{\scriptsize{\url{https://gym.openai.com/envs/\#classic\_control}}} to obtain rewardless state problems: Cartpole, Acrobot and MountainCar. The modifications can be summarized as follows: the state space of each problem is discretized into bins. Whenever a problem is initialized, a predefined percentage of these bins is marked as rewardless states. If, while learning, an agent reaches one of those states, no feedback from the environment is given back to the agent. In our experiments, we tried setting the percentage of rewardless states to 0\%, 10\%, 20\%, 30\%, 40\% and 50\%. An example of a state space with rewardless states is shown in Figure \ref{fig:implementation:rewardless_states}.
    In our experiments, we chose three RL algorithms: Q-learning, PPO and DQN. In addition, for the EA-only case, we chose genetic programming (GP) as it is also used in our implementation of Evo-RL (or goal is not, indeed, to compare the performance of different EAs). For all algorithms, we set the computational budget to 60,000 evaluations. 
    In \emph{Evo-RL}, for the EA part, a population of 30 agents per generation is adopted and the number of generations is chosen to be 200. Each individual is given 10 episodes for training in the reinforcement learning phase. Therefore, the total budget is 60,000 evaluations (200 (generations) $\times$ 30 (individuals per generation) $\times$ 10 (learning episodes per individual) = 60,000 (evaluations).
    In the \emph{EA-only} case, 2,000 generations are used in order to reach the 60,000 evaluations (30 (generations) $\times$ 2000 (individual per generation)). In the \emph{RL-only} case, each agent is given 60,000 episodes for training.
   
    For both Evo-RL and EA-only cases, the adopted selection scheme is \emph{tournament selection} with 3 candidates, the crossover rate is 50\%, the mutation rate is 15\%, and the mutation rate for the inherited behavior is 20\% with an independent probability of 10\% mutating each element.
    Due to the stochastic nature of the experiments, each agent is evaluated 100 times and only the average performance is considered. All experiments are repeated $10$ times and all results are presented in the form of mean and Standard Error of the Mean (SEM). The SEM is defined as the standard deviation weighted by the square root of the number of observations/trials (in our case N = 10)~\cite{book:Cambridge_Dictionary_of_Statistics}:
     \begin{equation}
     \sigma_x^- = \frac{\sigma}{\sqrt{N}}, \quad N=10
     \end{equation}
    Figure \ref{fig:cartpole} shows the performance of Q-learning, PPO and DQN in the first row; the second row shows the performance of each of those algorithms when operating within Evo-RL (denoted by adding the prefix ``e-''). The red straight line shows the reward threshold, i.e., the value of reward to reach in order to consider the problem solved and thus terminate the experiment. For CartPole it is 195, and for Acrobot it is -100. As shown, the performance of all RL algorithms when operating alone deteriorate as the percentage of rewardless states increases. On the other hand, the Evo-RL version of each of those algorithms converges and the problem is solved in most of the cases within the given computational budget.
   
    Furthermore, Figure \ref{fig:evaluation:CartPole:EA-PPO:multi} shows the ratio of instinctive behaviour relative to the learned behaviour for the CartPole problem when adopting e-Q learning. As shown in the figure, the instinctive behaviour increases as the ratio of rewardless states increases.
    
    Figure\ref{fig:EA_only} shows the performance of the EA-only case for the three problems. Finally, Tables \ref{tab:evaluation} and \ref{tab:evaluation2} summarize the final rewards after the last evaluation for all problems and algorithms. The number after the ``@'' denotes the number of evaluations needed for solving the problem. {\color{blue} Blue} indicates that the problem is completely solved and {\color{red} red} indicates that it was not solved.
    Of note, some runs of Evo-RL missed the reward threshold with a small margin and, based on the convergence rate, they would have likely solved the problem if given more evaluations.

 %%%%%%%%%%%%%%%%%%%%%%%%%%%%%%%%%%% Table 1 %%%%%%%%%%%%%%%%%%%%%%%%%%%%%%%%%
 \begin{table*}[!ht]
    \caption{Final rewards after 60,000 evaluations (mean and SEM over 10 trials). Top: EA-only, Q-Learning and eQ-learning (ours). Down: PPO, ePPO (ours), DQN and eDQN (ours). The number after the "@" denotes the number of evaluations needed for solving the problem. {\color{blue} Blue} indicates that the problem is solved and {\color{red} red} indicates the opposite.
            }
        \centering
        %\resizebox{\textwidth}{!}{%
        \begin{tabular}{p{2.1cm}  p{1.5cm} c  c c }
            \toprule
                {Environment} & Rewardless States [\%] & {\large EA-Only} & {\large Q-learning} & {\large eQ-learning (ours)} \\
            \midrule
                CartPole-v0
                & 0\%  
                &  {\color{red} \textbf{140.9} {\footnotesize $\pm 20.2$}}      
                & {\color{blue} \textbf{195.3} {\footnotesize $\pm 0.1$ @ 910}}    
                & {\color{blue} \textbf{196.3} {\footnotesize $\pm 0.3$ @ 10,800}}  
                \\ 
                & 10\% 
                & {\color{red} \textbf{101.3} {\footnotesize $\pm 18.0$}}      
                & {\color{red} \textbf{118.4} {\footnotesize $\pm 25.9$}}         
                & {\color{blue} \textbf{196.0} {\footnotesize $\pm 0.4$ @ 51,000}}       
                \\ 
                & 20\% 
                & {\color{red}\textbf{115.1} {\footnotesize $\pm 19.8$}}      
                & {\color{red} \textbf{\hspace{0.25cm}78.8}  {\footnotesize $\pm 24.1$}} 
                & {\color{blue} \textbf{196.2} {\footnotesize $\pm 0.3$ @ 31,200}}       
                \\ 
                & 30\% 
                &  {\color{red} \textbf{155.1} {\footnotesize $\pm 17.0$}}      
                & {\color{red}  \textbf{\hspace{0.25cm}63.9}  {\footnotesize $\pm 21.0$}} & {\color{blue} \textbf{196.2} {\footnotesize $\pm 0.4$ @ 31,200}}       
                \\ 
                & 40\% 
                & N.A.              
                & {\color{red} \textbf{\hspace{0.25cm}42.2}  {\footnotesize $\pm 16.4$}}
                & {\color{red}  \textbf{194.8} {\footnotesize $\pm 1.0$}}       
                \\ 
                & 50\% 
                & N.A.                     
                & {\color{red} \textbf{\hspace{0.25cm}41.8}  {\footnotesize $\pm 16.6$}} 
                & {\color{red} \textbf{192.0} {\footnotesize $\pm 3.2$}}         
                \\
                \hline
    
                % \multirow{2}{*}{{\large Acrobot-v1}} & 
                Acrobot-v1 
                & 0\%  
                &  {\color{red} \textbf{-109.5} {\footnotesize $\pm 6.5$}}           
                &  {\color{red} \textbf{-249.6} {\footnotesize $\pm 12.5$}}      
                & {\color{blue} \textbf{-88.6} {\footnotesize $\pm 2.2$ @ 43,500}}    
                \\ 
                & 10\% 
                &  {\color{red} \textbf{-106.5} {\footnotesize $\pm 5.9$}}           
                &  {\color{red}          \textbf{-298.6} {\footnotesize $\pm 32.9$}}     
                &  {\color{blue} \textbf{-96.3} {\footnotesize $\pm 1.2$ @ 43,200}}    
                \\ 
                & 20\% 
                &  {\color{red}  \textbf{-108.2} {\footnotesize $\pm 5.3$}} 
                &  {\color{red} \textbf{-344.9} {\footnotesize $\pm 40.4$}}      
                &  {\color{blue} \textbf{-96.6} {\footnotesize $\pm 1.0$ @ 33,900}}    
                \\ 
                & 30\% 
                & {\color{blue} \textbf{\hspace{0.25cm}-99.1}  {\footnotesize $\pm 5.8$ @ 10,860}}
                & {\color{red} \textbf{-344.8} {\footnotesize $\pm 40.9$}}      
                & {\color{blue} \textbf{-98.9} {\footnotesize $\pm 2.6$ @ 55,800}}    
                \\ 
                & 40\% 
                & N.A.                             
                &  {\color{red} \textbf{-430.3} {\footnotesize $\pm 33.7$}}      
                &  {\color{blue} \textbf{-97.0} {\footnotesize $\pm 2.3$ @ 49,800}}    
                \\ 
                & 50\% 
                & N.A.                  
                &  {\color{red} \textbf{-464.8} {\footnotesize $\pm 25.8$}}     
                &  {\color{blue} \textbf{-99.6} {\footnotesize $\pm 2.4$ @ 56,700}}    
                \\
                \hline
        \end{tabular}
        %}
        \label{tab:evaluation}
    \end{table*}
  %%%%%%%%%%%%%%%%%%%%%%%%%%%%%%%%%%% Table 2 %%%%%%%%%%%%%%%%%%%%%%%%%%%%%%%%%
    \begin{table*}[!ht]
        \caption{}
        \centering
        \begin{tabular}{ p{1.5cm}  p{1cm}  c c  c c}
            \toprule
                {Environment} &  & {\large PPO} & {\large ePPO (ours)} & {\large DQN} & {\large eDQN (ours)} \\
            \midrule
                CartPole &
                    0\%  
                    &  {\color{blue} \textbf{195.2} {\footnotesize $\pm 0.0$ @ 4,370}}         
                    & {\color{blue} \textbf{196.9} {\footnotesize $\pm 0.3$ @ 14,400}}  
                    &   {\color{blue} \textbf{195.5} {\footnotesize $\pm 0.1$ @ 160}}     
                    &   {\color{blue} \textbf{198.8} {\footnotesize $\pm 0.6$ @ 1200}}    \\ 
                    & 10\% 
                    &  {\color{red} \textbf{151.9}  {\footnotesize $\pm 20.9$}}         
                    & {\color{blue} \textbf{197.0} {\footnotesize $\pm 0.5$ @ 56,100}}  
                    & {\color{blue} \textbf{195.5} {\footnotesize $\pm 0.1$ @ 880}}     
                    & {\color{blue} \textbf{199.5} {\footnotesize $\pm 0.4$ @ 900}}     
                    \\ 
                    & 20\% 
                    &  {\color{red} \textbf{125.6} {\footnotesize $\pm 18.9$}}       
                    & {\color{blue} \textbf{196.8} {\footnotesize $\pm 0.5$ @ 51,900}}  
                    & {\color{blue} \textbf{195.6} {\footnotesize $\pm 0.1$ @ 3680}}    
                    & {\color{blue} \textbf{199.5} {\footnotesize $\pm 0.2$ @ 900}}     
                    \\ 
                    & 30\% 
                    &  {\color{red} \textbf{114.6} {\footnotesize $\pm 18.8$}}      
                    & {\color{blue} \textbf{196.1} {\footnotesize $\pm 0.2$ @ 51,600}}  
                    &   {\color{red} \textbf{142.9} {\footnotesize $\pm 25.4$}}         
                    &   {\color{blue} \textbf{199.2} {\footnotesize $\pm 0.5$ @ 2100}}    \\ 
                    & 40\% 
                    & {\color{red} \textbf{112.6} {\footnotesize $\pm 16.5$}}
                    & {\color{blue} \textbf{198.0} {\footnotesize $\pm 0.5$ @ 47,100}}  
                    & {\color{red} \textbf{139.6} {\footnotesize $\pm 26.9$}}         
                    & {\color{blue} \textbf{198.1} {\footnotesize $\pm 0.6$ @ 32100}}  
                    \\ 
                    & 50\% 
                    &  {\color{red} \textbf{\hspace{0.25cm}81.0}  {\footnotesize $\pm 20.5$}}                                                             
                    & {\color{blue} \textbf{196.9} {\footnotesize $\pm 0.3$ @ 31,500}}  
                    &   {\color{red} \textbf{121.0} {\footnotesize $\pm 28.8$}}         
                    &   {\color{blue} \textbf{198.5} {\footnotesize $\pm 0.6$ @ 45600}}   \\
                \hline
                Acrobot
                & 0\%  
                &  {\color{blue} \textbf{\hspace{0.25cm}-99.0}  {\footnotesize $\pm 0.2$ @ 12,300}}          
                &    {\color{blue} \textbf{\hspace{0.25cm}-99.0}  {\footnotesize $\pm 0.2$ @ 12,300}}          
                &  {\color{blue} \textbf{\hspace{0.25cm}-99.7}   {\footnotesize $\pm 0.1$ @ 1,270}}        
                &   {\color{blue} \textbf{-95.8} {\footnotesize $\pm 0.8$ @ 1,500}}     
                \\ 
                & 10\% 
                &  {\color{red} \textbf{-179.1} {\footnotesize $\pm 50.7$}} 
                &  {\color{blue} \textbf{\hspace{0.25cm}-97.3}  {\footnotesize $\pm 1.0$ @ 47,100}}          
                &  {\color{red} \textbf{-118.0} {\footnotesize $\pm 17.7$}}               &  {\color{blue} \textbf{-94.1} {\footnotesize $\pm 1.3$ @ 2,400}}     
                \\ 
                & 20\% 
                &  {\color{red} \textbf{-259.3} {\footnotesize $\pm 61.9$}}  
                &  {\color{blue} \textbf{\hspace{0.25cm}-97.2}  {\footnotesize $\pm 0.9$ @ 5,700}}           
                &  {\color{blue} \textbf{\hspace{0.25cm}-99.8}   {\footnotesize $\pm 0.1$ @ 13,120}}       
                &   {\color{blue} \textbf{-96.3} {\footnotesize $\pm 0.9$ @ 1,500}}     \\ 
                & 30\% 
                &   {\color{red} \textbf{-299.7} {\footnotesize $\pm 63.3$}}                          
                &    {\color{blue} \textbf{\hspace{0.25cm}-98.3}  {\footnotesize $\pm 0.4$ @ 15,000}}          
                &  {\color{blue} \textbf{\hspace{0.25cm}-99.7}   {\footnotesize $\pm 0.1$ @ 19,340}}       
                &   {\color{blue} \textbf{-90.3} {\footnotesize $\pm 2.1$ @ 7,800}}     
                \\ 
                & 40\% 
                & {\color{red} \textbf{-353.6} {\footnotesize $\pm 52.4$}}                    
                & {\color{blue} \textbf{\hspace{0.25cm}-97.3}  {\footnotesize $\pm 0.7$ @ 50,400}}          
                &  {\color{red} \textbf{-139.5} {\footnotesize $\pm 38.0$}} 
                &  {\color{blue} \textbf{-93.1} {\footnotesize $\pm 1.3$ @ 6,300}}     
                \\ 
                & 50\% 
                & {\color{red} \textbf{-427.5} {\footnotesize $\pm 46.2$}}                           
                &  {\color{red} \textbf{-101.2} {\footnotesize $\pm 5.5$}}
                &  {\color{red} \textbf{-175.6} {\footnotesize $\pm 45.9$}}    
                &  {\color{blue} \textbf{-90.8} {\footnotesize $\pm 1.5$ @ 6,600}}     
                \\
                \hline
                % \multirow{2}{*}{{\large MountainCar-v0}} & 
                MountainCar
                & 0\%  
                &   {\color{red} \textbf{-200.0} {\footnotesize $\pm 0.0$}}
                &   {\color{red} \textbf{-136.5} {\footnotesize $\pm 1.3$}}   
                &   {\color{red} \textbf{-189.3} {\footnotesize $\pm 6.8$}} 
                &   {\color{blue} \textbf{-106.8} {\footnotesize $\pm 0.6$ @ 17,100}}    \\ 
                & 10\% 
                &   {\color{red} \textbf{-200.0} {\footnotesize $\pm 0.0$}}   
                &   {\color{red} \textbf{-133.5} {\footnotesize $\pm 1.4$}}    
                &   {\color{red} \textbf{-192.0} {\footnotesize $\pm 5.1$}} 
                &   {\color{blue} \textbf{-107.0} {\footnotesize $\pm 0.7$ @ 19,500}}    \\ 
                & 20\% 
                &   {\color{red} \textbf{-197.3} {\footnotesize $\pm 2.2$}}    
                &   {\color{red} \textbf{-131.5} {\footnotesize $\pm 1.0$}}    
                &   {\color{red} \textbf{-200.0} {\footnotesize $\pm 0.0$}} 
                &   {\color{blue} \textbf{-107.8} {\footnotesize $\pm 0.5$ @ 57,600}}    \\ 
                & 30\% 
                &  {\color{red} \textbf{-195.6} {\footnotesize $\pm 2.5$}}    
                &   {\color{red} \textbf{-133.0} {\footnotesize $\pm 1.5$}}    
                &   {\color{red} \textbf{-199.1} {\footnotesize $\pm 0.9$}}  
                &   {\color{red} \textbf{-110.0} {\footnotesize $\pm 1.6$}}    
                \\
            \bottomrule
        \end{tabular}
        \label{tab:evaluation2}
    \end{table*}
\section{Discussion and Future work}
The presented hybrid approach REAL outperforms EA only and RL only, even when adopting state-of-the-art RL algorithms such as PPO and DQN. Our approach has a number of points of strength: it integrates RL in an EA framework, hence benefiting from both methodologies. Furthermore, it works with any RL algorithm and can handle problems where the reward function is not valid in all states. 

For future work, we acknowledge the importance of applying this algorithm on real-world problems where the state space has rewardless states, e.g., due to the ambiguity of the environment or to the complexity of defining a reward function suited for the majority of states in that environment. Further, we will adapt our work for solving meta-learning problems, as achieved for instance in \cite{finn2017model}, \cite{wang2016learning}, \cite{wang2019paired} and \cite{li2017meta}.

\newpage
\bibliographystyle{plainnat}
\bibliography{eRL}

\end{document}